\pdfoutput=1
\documentclass[11pt]{article}
\usepackage[final]{acl}
\usepackage{times}
\usepackage{latexsym}
\usepackage[T1]{fontenc}
\usepackage[utf8]{inputenc}
\usepackage{microtype}
\usepackage{inconsolata}
\usepackage{graphicx}
\usepackage{subcaption}
\usepackage[whole]{bxcjkjatype}
\usepackage{booktabs}
\usepackage{amsmath,amssymb,amsthm}
\usepackage{bm}
\usepackage{mathtools}
\usepackage{algorithmic}
\usepackage{algorithm}
\usepackage{color}
\usepackage{multirow}
\usepackage{xurl}
\usepackage{xcolor}
\usepackage{xspace}

\definecolor{myblue}{RGB}{65,117,181}
\definecolor{myorange}{RGB}{183,138,45}

\newcommand{\tokA}{\textcolor{myblue}{$\bullet$}}
\newcommand{\tokB}{\textcolor{myorange}{$\bullet$}}
\newcommand{\meanA}{\textcolor{myblue}{$\bigstar$}}
\newcommand{\meanB}{\textcolor{myorange}{$\bigstar$}}

\newcommand{\OurTool}{SOCM\xspace}
\newcommand{\MU}[1]{\bm{\mu}(#1)}
\newcommand{\SIGMA}[1]{\bm{\Sigma}(#1)}

\newtheorem{theorem}{Theorem}

\newtheorem{definition}{Definition}
\newtheorem{assumption}{Assumption}

\theoremstyle{remark}

\theoremstyle{remark}
\newtheorem*{intuition}{Intuition}

\newcommand{\GTEsmall}{GTE$_\text{small}$\xspace}
\newcommand{\GTEbase}{GTE$_\text{base}$\xspace}
\newcommand{\EFivesmall}{E5$_\text{small}$\xspace}
\newcommand{\EFivebase}{E5$_\text{base}$\xspace}

\title{Why Mean Pooling Works:\\ Quantifying Second-Order Collapse in Text Embeddings}

\author{
 \textbf{Tomomasa Hara\textsuperscript{1}},
 \textbf{Hiroto Kurita\textsuperscript{1}},
 \textbf{Masaaki Imaizumi\textsuperscript{2,3,4}},
 \textbf{Kentaro Inui\textsuperscript{5,1,4}},
 \textbf{Sho Yokoi\textsuperscript{6,1,4}}
\\
 \textsuperscript{1}Tohoku University,
 \textsuperscript{2}The University of Tokyo,
 \textsuperscript{3}Kyoto University,\\
 \textsuperscript{4}RIKEN,
 \textsuperscript{5}MBZUAI,
 \textsuperscript{6}NINJAL
\\
 \texttt{\{hara.tomomasa.s8, hiroto.kurita.q4\}@dc.tohoku.ac.jp}\\
 \texttt{imaizumi@g.ecc.u-tokyo.ac.jp}\quad \texttt{kentaro.inui@mbzuai.ac.ae}\quad \texttt{yokoi@ninjal.ac.jp}
}

\begin{document}
\maketitle

\begin{abstract}
For constructing text embeddings, mean pooling, which averages token embeddings, is the standard approach.
This paper examines whether mean pooling actually works well in real models.
First, we note that mean pooling can collapse information beyond the first-order statistics of the token embeddings, such as second-order statistics that capture their spatial structure, potentially mapping distinct token embedding distributions to similar text embeddings.
Motivated by this concern, we propose a simple metric to quantify such a collapse induced by mean pooling.
Then, using this metric, we empirically measure how often this collapse occurs in actual models and texts, and find that modern text encoders are robust to this collapse.
In particular, contrastive fine-tuned text encoders tend to be less prone to the collapse than their pretrained backbone models.
We also find that the robustness of these text encoders lies in the concentration of token embeddings within each text.
In addition, we find that robustness to the collapse, as quantified by our proposed metric, correlates with downstream task performance.
Overall, our findings offer a new perspective on why modern text encoders remain effective despite relying on seemingly coarse mean pooling.
\end{abstract}

\section{Introduction} \label{sec:introduction}
\begin{figure}[t]  
    \centering
    \includegraphics[width=\linewidth]{Figure/image/overview.pdf}
    \caption{
        Overview of this work. 
        \textbf{Top:} Mean pooling can map distinct token embedding distributions to similar text embeddings.
        This is because mean pooling summarizes distributions using only their first-order statistics, collapsing higher-order statistics.
        \textbf{Bottom:} We empirically find that modern fine-tuned text encoders are robust to such a collapse (\S~\ref{sec:experiment}).
        Each panel visualizes token and text embeddings for two texts, output by BERT~\cite{Devlin2019-mb} and GTE$_\text{base}$~\cite{Li2023-oy} via PCA projection without centering.
        This example was discovered using the metric in \S~\ref{sec:mean_pooling_collapse}.
    }
    \label{fig:overview}
\end{figure}

Text embeddings, which represent sentences, paragraphs, and documents as single vectors, are used across a wide range of NLP tasks~\cite{Enevoldsen2025-hc}, including information retrieval~\cite{Thakur2021-ac} and automatic evaluation~\cite{Rei2020-pf}.
Also, text embeddings have become essential for modern retrieval-augmented generation~\cite{Lewis2020-gr} applications.
Given their widespread use, understanding text embeddings remains an important open challenge, and prior work has approached this problem from the perspectives of geometry~\cite{Xiao2023-vu} and dimensionality~\cite{Takeshita2025-eg}.
This paper focuses on the aggregation method that constructs embeddings from token-level representations.

As such, a standard aggregation method is \textit{mean pooling}, which averages token embeddings.
This simple aggregation has offered empirical advantages across embedding methods, from classical static token embeddings~\cite{Wieting2015-fd, Shen2018-gg} to modern contextualized embeddings from Transformer~\cite{Vaswani2017-kx} encoders~\cite{Reimers2019-gc}.
Indeed, some state-of-the-art text encoders use mean pooling~\cite{Lee2025-yl, Nussbaum2025-ys}.

While mean pooling is standard, does it actually work well in real models?
Figure~\ref{fig:overview} (top) illustrates a potential collapse mode:
even when two texts yield distinct token embeddings in the embedding space
([\tokA$,\cdots,$\tokA] $\neq$ [\tokB$,\cdots,$\tokB]), mean pooled text embeddings may become indistinguishable (\meanA\ $\approx$ \meanB).
This is because the mean-pooled text embedding corresponds only to the first-order statistic (mean) of the token embedding distribution.
Thus, mean pooling does not capture the spatial structure of token embeddings, as reflected in second- and higher-order statistics, such as the covariance matrix.
Turning to text embeddings via mean pooling, several prior works have proposed alternative approaches to mean-pooled text embeddings that represent a text as a list of token embeddings prior to mean pooling, such as optimal transport between token embedding lists~\cite{Kusner2015-ol, Zhao2019-mv, Yokoi2020-qw, Lee2022-lo} (\S~\ref{subsec:beyond_mean_pooling}).
Nevertheless, mean-pooled text embeddings remain dominant in practice, as they are both computationally efficient and empirically effective.

This potential collapse raises the question: do actual models suffer from it, or do they avoid it?
To address this question, we first propose a framework to evaluate the robustness of a text encoder to such a collapse.
Specifically, we introduce the degree of \textbf{S}econd-\textbf{O}rder \textbf{C}ollapse by \textbf{M}ean pooling, a metric that quantifies the severity of the second-order statistics collapse by mean pooling (\S~\ref{sec:mean_pooling_collapse}).

Next, using this proposed metric, we evaluate how well mean pooling works in actual models and texts.
Our empirical results suggest that mean pooling works well in modern text encoders.
In particular, we make the following findings:
(i) We find that contrastive fine-tuned Transformer text encoders tend to be less prone to the collapse than their pretrained backbone models, suggesting that these fine-tuned encoders are robust to the collapse (\S~\ref{sec:experiment}, Figure~\ref{fig:overview} bottom).
(ii) We also find that this robustness lies in the concentration of token embeddings within each text, which arises through the Transformer layer (\S~\ref{sec:how_avoid_collapse}).
(iii) We further observe that the proposed metric correlates with downstream task performance, suggesting that robustness to the collapse can be one factor in the success of modern text encoders on downstream tasks (\S~\ref{sec:correlation_downstream_task}).

Overall, by rethinking mean pooling, which may appear to be a coarse aggregation method, this paper offers a new perspective on the effectiveness of modern text encoders.

\section{Related Work} \label{sec:related_work}
This paper focuses on the coarseness of mean pooling in modern text embeddings, namely, its loss of higher-order statistics of token embeddings.
Text encoders with mean pooling have become a standard paradigm (\S~\ref{subsec:text_embeddings_mean_pooling}).
Meanwhile, as an alternative paradigm, methods that preserve higher-order statistics without pooling have also been proposed (\S~\ref{subsec:beyond_mean_pooling}); yet modern text encoders achieve empirical performance competitive with these methods.
We offer one perspective on why modern text encoders work well by rethinking mean pooling.

\subsection{Text Embeddings via Mean Pooling}\label{subsec:text_embeddings_mean_pooling}
In recent NLP, mean pooling over token embeddings from contrastive fine-tuned Transformer encoders has become a widely used paradigm for text embeddings~\cite{Reimers2019-gc, Gao2021-ds,  Wang2022-gw, Li2023-oy, Nussbaum2025-ys, Lee2025-yl}. 
This popularity likely reflects its compatibility with contrastive fine-tuning~\cite{Gao2021-ds} and computational efficiency, including low memory use~\cite{T2024-uo} and compatibility with approximate nearest neighbor search~\cite{Douze2025-wj}. 

Beyond these engineering motivations, this paper offers a new perspective on mean pooling effectiveness, highlighting that the collapse of higher-order statistics is limited in modern text encoders.

\subsection{Text Representations Preserving Higher-Order Statistics}\label{subsec:beyond_mean_pooling}
As an alternative paradigm to mean pooling, methods that handle text representations while preserving higher-order statistics have also been proposed.
One approach is to work directly with the list of token embeddings before pooling.
Distances of such lists have been computed using optimal transport, which measures the minimum cost of transporting one list to another~\cite{Kusner2015-ol, Zhao2019-mv, Yokoi2020-qw, Lee2022-lo}.
Similarly, ColBERT~\cite{Khattab2020-vk, Santhanam2022-ex} measured the distance of token embedding lists by matching each query token to its most similar document token and aggregating the resulting score.
BERTScore~\cite{Zhang2019-fv} also aggregated token-level similarities into an F-score-like metric.
As another direction, Sen2Pro~\cite{Shen2023-zb} represents each text as a probability distribution over text embeddings by sampling multiple embeddings via MC dropout and data augmentation.
GaussCSE~\cite{Yoda2024-xu} directly predicts first- and second-order statistics
to represent each text as a Gaussian distribution.
 
Meanwhile, modern text encoders with mean pooling are not only computationally lighter than these methods, but also achieve competitive empirical performance~\cite{Lee2022-lo}.
This paper investigates why text encoders remain effective even when representing texts via simple mean pooling.

\section{Preliminaries} \label{sec:preliminaries}
In this paper, we quantify how mean pooling, by using only the first-order statistics of the original token embeddings, collapses its second-order statistics (\S~\ref{sec:mean_pooling_collapse}).
In this section, we formally introduce token embeddings and these statistics.

We consider constructing text embeddings for two texts $t_1$ and $t_2$ using a model $\bm{f}$.
Given $t_i$, $\bm{f}$ outputs a token embedding list $\bm{X}_i$:
\begin{align}
    \bm{X}_i := [\bm{x}_{i,1}, \cdots, \bm{x}_{i,n_i}\,]=\bm{f}(t_i) \in \mathbb{R}^{d \times n_i}.
\end{align}
Here, $\bm{x}_{i,j} \in \mathbb{R}^d$ is the $j$-th token embedding in $t_i$, $d$ is the embedding dimension, and $n_i$ is the number of tokens in $t_i$.

Mean pooling averages token embeddings to construct a text embedding $\MU{\bm{X}_i} \in \mathbb{R}^d$:
\begin{equation}
    \MU{\bm{X}_i} := \frac{1}{n_i}\sum_{j=1}^{n_i}\bm{x}_{i,j}.
\end{equation}
This text embedding $\MU{\bm{X}_i}$ corresponds to the first-order statistic of $\bm{X}_i$ when viewed as an empirical distribution in the embedding space.

Meanwhile, the token embedding distribution also has a second-order statistic $\SIGMA{\bm{X}_i} \in \mathbb{R}^{d \times d}$:
\begin{align}
    &\SIGMA{\bm{X}_i} \\ \nonumber
    &:= \frac{1}{n_i}\sum_{j=1}^{n_i}
    (\bm{x}_{i,j} - \MU{\bm{X}_i})
    (\bm{x}_{i,j} - \MU{\bm{X}_i})^\top.
\end{align}
This second-order statistic captures the spatial structure of token embeddings.
For simplicity, we focus on it as the lowest-order statistic not retained by mean pooling.

Based on these definitions, we quantify the collapse induced by mean pooling (\S~\ref{sec:mean_pooling_collapse}) and examine how often it arises in practice (\S~\ref{sec:experiment}).

\section{Second-Order Collapse by Mean Pooling} \label{sec:mean_pooling_collapse}
In this section, we provide both intuitive and formal characterizations of collapse by mean pooling, based on first- and second-order statistics.
We first provide an intuitive understanding of when this collapse occurs by examining the similarity of first- and second-order statistics (\S~\ref{subsec:when_collapse}).
Based on this intuition, we then introduce a metric to quantify the severity of this collapse (\S~\ref{subsec:proposed_metric_socm}).

\subsection{When Does Collapse Arise?}\label{subsec:when_collapse}
\begin{figure}[t]
    \centering
    \includegraphics[width=0.9\columnwidth]{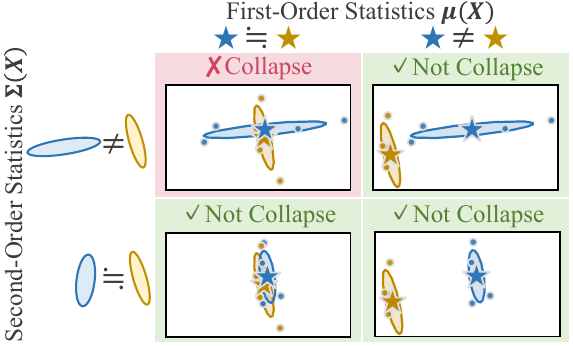}
    \caption{
        When mean pooling collapse arises (red) and does not (green).
        Mean pooling collapse occurs when the first-order statistics are similar while the second-order statistics differ.
        Each panel depicts a conceptual distribution in a two-dimensional embedding space, characterized by its first- and second-order statistics.
    }
    \label{fig:collapse_matrix}
\end{figure}

As discussed in \S~\ref{sec:introduction}, mean pooling may collapse two distinct token embedding distributions into similar text embeddings.
We characterize whether this collapse arises based on the similarity of their first-order statistics and that of their second-order statistics, as shown in Figure~\ref{fig:collapse_matrix}.

\paragraph{When Collapse Arises}
The collapse arises when first-order statistics are similar but second-order ones differ, i.e., $\MU{\bm{X}_1} \approx \MU{\bm{X}_2} \land \SIGMA{\bm{X}_1} \neq \SIGMA{\bm{X}_2}$.
In this case, distributions that are distinct up to the second-order statistics may collapse into similar representations after mean pooling.

\paragraph{When Collapse Does Not Arise}
We also consider when the collapse does not arise.
One case is when first-order statistics differ, i.e., $\MU{\bm{X}_1} \neq \MU{\bm{X}_2}$.
Here, the distributions are separated by their means unless the variance is excessively large.
Another case is when both first- and second-order statistics are similar, i.e., $\MU{\bm{X}_1} \approx \MU{\bm{X}_2}$ and $\SIGMA{\bm{X}_1} \approx \SIGMA{\bm{X}_2}$.
Here, the distributions are similar, as reflected in their means.

\subsection{Quantify the Severity of Collapse}
\label{subsec:proposed_metric_socm}
Based on these intuitions, we introduce the degree of \textbf{S}econd-\textbf{O}rder \textbf{C}ollapse by \textbf{M}ean pooling (hereafter referred to as \textbf{\OurTool}), a metric that quantifies the collapse of second-order statistics by mean pooling.
As discussed in \S~\ref{subsec:when_collapse}, whether the collapse by mean pooling arises depends on the similarity of first-order statistics and the similarity of second-order statistics.
Thus, we define \OurTool using the distance of the first-order statistics and the distance of the second-order statistics (\S~\ref{subsubsec:socm_definition}).
Specifically, we decompose the distance of token embedding distributions into first- and second-order components and construct \OurTool from them (\S~\ref{subsubsec:decomposition}).
We also design \OurTool to satisfy desirable properties such as boundary conditions and monotonicity (\S~\ref{subsubsec:properties}).

\subsubsection{Definition of \OurTool}\label{subsubsec:socm_definition}
Given two token embedding lists $\bm{X}_1$ and $\bm{X}_2$, \OurTool quantifies the severity of the collapse when applying mean pooling to them.
Specifically, \OurTool is defined as:
\begin{equation}\label{eq:socm}
    \mathrm{\OurTool}(d_\mu, d_\Sigma) := (1 - d_\mu)d_\Sigma.
\end{equation}
Here, $d_\mu$ is the distance of the first-order statistics $\MU{\bm{X}_1}$ and $\MU{\bm{X}_2}$, and $d_\Sigma$ is the distance of the second-order statistics $\SIGMA{\bm{X}_1}$ and $\SIGMA{\bm{X}_2}$.

$d_\mu$ is the scaled squared Euclidean distance:
\begin{equation}
    d_\mu:=\|\MU{\bm{X}_1} - \MU{\bm{X}_2} \|_2^2/4.
\end{equation}
We assume unit-norm means, $\|\MU{\bm{X}_i}\|_2=1$, as is common in the use cases of text embeddings~\cite{Enevoldsen2025-hc}.
Under this normalization, $d_\mu \in [0,1]$\footnote{A factor $1/4$ rescales $d_\mu$ to $[0,1]$.} and corresponds to a common distance for text embeddings~\cite{Enevoldsen2025-hc}.

$d_\Sigma$ is defined as the scaled Bures Wasserstein distance~\cite{Dowson1982-sq}:
\begin{align}
    &d_\Sigma:= \mathrm{tr}\bigl(
           \SIGMA{\bm{X}_1} + \SIGMA{\bm{X}_2} \nonumber\\
    &-2\bigl(\SIGMA{\bm{X}_1}^{1/2}\SIGMA{\bm{X}_2}\SIGMA{\bm{X}_1}^{1/2}\bigr)^{1/2}
       \bigr)/4.
\end{align}
We assume $\mathrm{tr}(\SIGMA{\bm{X}_i})\leq 2$ to put $d_\Sigma$ in the same range as $d_\mu$.
This assumption corresponds to the scenario described in \S~\ref{subsec:when_collapse}, where the variance of token embeddings does not become excessively large.
Under this assumption, $d_\Sigma \in [0,1]$\footnote{As with $d_\mu$, the factor $1/4$ rescales $d_\Sigma$ to $[0,1]$ under the trace bound.}, matching the range of $d_\mu$.

Since $d_\mu, d_\Sigma \in [0,1]$ as above, Eq.~\eqref{eq:socm} implies $\mathrm{\OurTool} \in[0,1]$.
This \OurTool value indicates a more severe collapse for larger values.
As discussed in \S~\ref{subsec:when_collapse}, the collapse arises when first-order statistics are similar ($d_\mu$ is small),  but second-order statistics differ ($d_\Sigma$ is large).
\OurTool reflects this intuition via $(1-d_\mu)d_\Sigma$.

In the following subsections, we discuss the validity of this \OurTool design.

\subsubsection{Decomposing Distributional Distance} \label{subsubsec:decomposition}
We adopt $d_\mu$ and $d_\Sigma$ as the first- and second-order distances, respectively.
This choice follows from decomposing the distance of token embedding distributions into first- and second-order components.

\paragraph{Decomposition}
$d_\mu$ and $d_\Sigma$ correspond to the first- and second-order components of the $L_2$-Wasserstein distance $W_2^2$ of token embedding distributions~\cite{Dowson1982-sq}:
\begin{align}
    &W_2^2\bigl(\mathcal{N}(\MU{\bm{X}_1},\SIGMA{\bm{X}_1}),
                \mathcal{N}(\MU{\bm{X}_2},\SIGMA{\bm{X}_2})\bigr)/4 \nonumber\\
    &= d_\mu + d_\Sigma.
\end{align}
For simplicity, we characterize each token embedding list $\bm{X}_i$ as a Gaussian $\mathcal{N}(\MU{\bm{X}_i},\SIGMA{\bm{X}_i})$.
This means that higher-order moments are not explicitly distinguished, though the Gaussian characterization is motivated by computational tractability and stability in Wasserstein-based evaluation (see Appendix~\ref{sec:gaussian_characterization} for details).

\paragraph{Motivation for $W_2^2$} 
This decomposition of the Wasserstein distance is a natural choice for quantifying the severity of the collapse by mean pooling. 
The Wasserstein distance is a widely used metric for measuring the distance between distributions, considering not only first- but also higher-order statistics~\cite{Villani2009-go}.
Indeed, optimal transport, which measures semantic similarity over token embedding lists (discussed in \S~\ref{subsec:beyond_mean_pooling}), is based on this Wasserstein distance~\cite{Peyre2018-yj}. 
Therefore, $d_\mu$ and $d_\Sigma$, which decompose the Wasserstein distance, are natural components for constructing \OurTool.

\subsubsection{Validity of \OurTool Form} \label{subsubsec:properties}
We adopt Eq.~\eqref{eq:socm} as \OurTool.
To verify its validity, we enumerate desirable properties for quantifying the severity of the collapse and show that this form satisfies all of them.

\paragraph{Desirable Properties}
To quantify the collapse severity, we define five desirable properties:
\begin{itemize}
    \item[(a)] When Collapse Arises: \\ 
    $\displaystyle{d_\mu = 0 \land d_\Sigma = 1 \Leftrightarrow \mathrm{\OurTool} = 1}$.
    \item[(b)] When Collapse Does Not Arise: \\
    $d_\mu = 1 \lor d_\Sigma = 0 \Leftrightarrow \mathrm{\OurTool} = 0$.
    \item[(c)] Monotonicity in $d_\mu$: $\displaystyle{\frac{\partial\,\mathrm{\OurTool}}{\partial d_\mu} \leq 0}$.
    \item[(d)] Monotonicity in $d_\Sigma$: $\displaystyle{\frac{\partial\,\mathrm{\OurTool}}{\partial d_\Sigma} \geq 0}$.
    \item[(e)] Interaction of $d_\mu$ and $d_\Sigma$: $\displaystyle{\frac{\partial^2\,\mathrm{\OurTool}}{\partial d_\mu\,\partial d_\Sigma} \leq 0}$.
\end{itemize}

\paragraph{(a) When Collapse Arises}
Property (a) requires that $\mathrm{\OurTool}=1$ if and only if the collapse arises in the extreme case: $d_\mu=0 \land d_\Sigma=1$.
As discussed in \S~\ref{subsec:when_collapse}, collapse arises when first-order statistics are similar but second-order statistics differ.
This extreme case corresponds to identical first-order statistics ($d_\mu=0$) and maximally different second-order statistics ($d_\Sigma=1$).

\paragraph{(b) When Collapse Does Not Arise}
Property (b) requires that $\mathrm{\OurTool}=0$ if and only if the collapse does not arise: $d_\mu=1 \lor d_\Sigma=0$.
As discussed in \S~\ref{subsec:when_collapse}, collapse does not arise when first-order statistics differ or when both first- and second-order statistics are similar.
$d_\mu=1 \lor d_\Sigma=0$ indicates these cases.
Specifically, $d_\mu=1$ indicates maximally different first-order statistics.
Additionally, $d_\Sigma=0$ indicates identical second-order statistics, so similarity is determined by $d_\mu$.

\paragraph{(c) Monotonicity in $d_\mu$}
Property (c) requires that \OurTool is monotonically non-increasing in $d_\mu$.
This reflects that the collapse is less severe as the first-order statistics become more distant.

\paragraph{(d) Monotonicity in $d_\Sigma$}
Property (d) requires that \OurTool is monotonically non-decreasing in $d_\Sigma$.
This reflects that the collapse becomes more severe as the second-order statistics become more distant.

\paragraph{(e) Interaction between $d_\mu$ and $d_\Sigma$}
Property (e) requires that the impact of $d_\Sigma$ on \OurTool decreases as $d_\mu$ increases.
While property (d) requires \OurTool to increase with $d_\Sigma$, this increase should be smaller for large $d_\mu$.
When first-order statistics are already distant (large $d_\mu$), they capture distributional differences, so second-order differences matter less.
The mixed derivative enforces this: as $d_\mu$ increases, \OurTool becomes less sensitive to $d_\Sigma$.

\paragraph{Form of \OurTool Satisfying the Properties}
\begin{figure}[t]  
    \centering
    \includegraphics[width=0.8\linewidth]{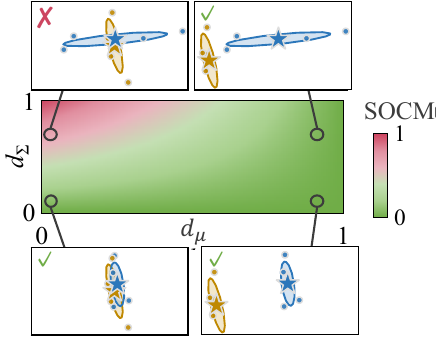}
    \caption{
    SOCM values for each combination of $(d_\mu, d_\Sigma)$.
    The figure confirms that SOCM satisfies all properties (a)–(e).
    The four corners correspond to the scenarios in Figure~\ref{fig:collapse_matrix}.
    SOCM is higher when mean pooling collapse occurs (top-left) and lower when it does not (the other three cases).
    }
    \label{fig:socm_property}
\end{figure}

The adopted form in Eq.~\eqref{eq:socm} satisfies all desirable properties (a)–(e) (see Appendix~\ref{sec:appendix_proof} for the proof).
Figure~\ref{fig:socm_property} plots values of \OurTool over $d_\mu$ and $d_\Sigma$, confirming properties (a)–(e).
The four corners of the plot correspond to the scenarios in Figure~\ref{fig:collapse_matrix}.
\OurTool is high when collapse arises and low otherwise.

\section{Experiment} \label{sec:experiment}
Using \OurTool, we empirically measured how often the collapse induced by mean pooling arose across actual models and texts.
We found that contrastive fine-tuned Transformer text encoders tend to be less prone to such collapse than their pretrained backbone models.

\subsection{Experimental Procedure and Setting}
\paragraph{Overview}
To examine how often the collapse arises, we prepared a text pair dataset $D=\{(t_{1}, t_{2})\}$ and a model $\bm{f}$.
Given a text $t_i$, we obtain a token embedding list $\bm{X}_i$ from the model $\bm{f}$.
We computed \OurTool for each token embedding list pair $(\bm{X}_{1}, \bm{X}_{2})$.

\paragraph{Normalization of Token Embedding Lists}
We normalized each token embedding list.
The definition of \OurTool in \S~\ref{subsubsec:socm_definition} assumes this normalization, $\|\MU{\bm{X}}\|_2=1$.
This assumption aligns with standard practice in text embedding applications, where a unit-normalized embedding $\MU{\bm{X}}/\|\MU{\bm{X}}\|_2$ is used~\cite{Enevoldsen2025-hc}.
To satisfy this assumption, given a token embedding list $\bm{X}_{i} = [\bm{x}_{i,1}, \ldots, \bm{x}_{i,n_{i}}] \in \mathbb{R}^{d \times n_{i}}$ from model $\bm{f}$, we use the normalized version $\bm{X}_{i}^\text{norm}$ defined as:
\begin{equation}\label{eq:normalize_token_embedding}
    \bm{X}_{i}^\text{norm} = \left[\frac{\bm{x}_{i,1}}{\|\MU{\bm{X}_{i}}\|_2}, \ldots, \frac{\bm{x}_{i,n_{i}}}{\|\MU{\bm{X}_{i}}\|_2}\right] \in \mathbb{R}^{d \times n_{i}}.
\end{equation}
$\MU{\bm{X}_{i}^\text{norm}} = \MU{\bm{X}_{i}}/\|\MU{\bm{X}_{i}}\|_2$ holds under this normalization (see Appendix~\ref{sec:appendix_normalization}).
We then computed \OurTool for each pair $(\bm{X}_{1}^\text{norm}, \bm{X}_{2}^\text{norm})$.

\paragraph{Dataset}
We constructed a dataset of text pairs from Wikipedia~\cite{Gao2021-ds}, which contains 1 million texts.
Specifically, we randomly sampled 1,000 texts and generated 499,500 text pairs by comparing them pairwise.
Experiments on another dataset, MS MARCO~\cite{Nguyen2018-ql}, led to similar conclusions (Appendix~\ref{sec:appendix_msmarco_results}).

\paragraph{Model}
We examined popular Transformer text encoders that use mean pooling.
Specifically, we used the following models, each obtained by contrastive fine-tuning the respective backbone model: Unsupervised SimCSE~\cite{Gao2021-ds} with mean pooling, \EFivebase~\cite{Wang2022-gw}, and \GTEbase~\cite{Li2023-oy} (BERT~\cite{Devlin2019-mb}); all-MiniLM-L12-v2~\cite{Reimers2019-gc}, \EFivesmall~\cite{Wang2022-gw}, and \GTEsmall~\cite{Li2023-oy} (MiniLM~\cite{Wang2020-mg}); all-mpnet-base-v2~\cite{Reimers2019-gc} (MPNet~\cite{Song2020-ls}); and nomic-embed-text-v1.5~\cite{Nussbaum2025-ys} (nomic-bert-2048~\cite{Nussbaum2025-ys}).
We also applied \OurTool to the respective backbone models.
 
\subsection{Results}
\paragraph{Quantitative Analysis}
\begin{table}[t]\centering
    \small
    \begin{tabular}{llr}
        \toprule
        \textbf{Model} &\textbf{Avg. \OurTool} $\downarrow$ \\
        \midrule
        $\text{BERT}$ &$0.396$ \\
        $\rightarrow \text{Unsup-SimCSE-mean}$ &$\bm{0.193}$ $(-0.203)$ \\
        $\rightarrow \text{E5}_\text{base}$ &$\bm{0.029}$ $(-0.367)$ \\
        $\rightarrow \text{GTE}_\text{base}$ &$\bm{0.018}$ $(-0.378)$ \\
        \midrule
        $\text{MiniLM}$ &$0.242$  \\
        $\rightarrow \text{all-MiniLM-L12-v2}$ &$0.313$ $\hspace{1.2mm}(+0.071)$ \\
        $\rightarrow \text{E5}_\text{small}$ &$\bm{0.099}$ $(-0.143)$ \\
        $\rightarrow \text{GTE}_\text{small}$ &$\bm{0.055}$ $(-0.187)$ \\
        \midrule
        $\text{MPNet}$ &$0.117$ \\
        $\rightarrow \text{all-mpnet-base-v2}$ &$\bm{0.100}$ $(-0.017)$ \\
        \midrule
        $\text{nomic-bert-2048}$ &$0.139$ \\
        $ \rightarrow \text{nomic-embed-text-v1.5}$ &$\bm{0.122}$ $(-0.017)$ \\
        \bottomrule
    \end{tabular}
    \caption{
        Average SOCM values for each model on text pairs from Wikipedia.
        For text encoders derived from backbone models, values in parentheses show the change in SOCM.
        Bold values indicate a reduction.
    }
    \label{tab:collapse_wiki}
\end{table}

Table~\ref{tab:collapse_wiki} shows the average \OurTool for each model on Wikipedia.
Overall, contrastive fine-tuned text encoders tended to have lower \OurTool than their backbone models.
This suggests that contrastive fine-tuned text encoders are less prone to collapse induced by mean pooling.

\paragraph{Qualitative Analysis}
We further validated our findings by visualizing embedding spaces.
The bottom of Figure~\ref{fig:overview} shows 2D PCA projections (without centering) for BERT and \GTEbase on two semantically unrelated texts: ``\textit{Virginia Woolf set many scenes of her novel ``Night and Day'' (1919) in Russell Square}'' and ``\textit{Ghiz was born in Charlottetown, Prince Edward Island, to Atallah Joseph Ghiz, a Lebanese corner store owner, and Marguerite F. Ghiz (née McKarris).}''.
For this example, BERT exhibits high $\mathrm{\OurTool} = 0.618$ while \GTEbase shows low $\mathrm{\OurTool} = 0.024$.\footnote{\OurTool values are computed in the original 768-dimensional space before dimensionality reduction for visualization.}
For BERT, the token embedding distributions differed in spread but had similar means.
In contrast, for \GTEbase, the means of the two token embedding distributions were separated.
These visualizations are consistent with the quantitative analysis.

In summary, these results indicate that contrastive fine-tuned text encoders are more robust to the collapse induced by mean pooling.
This suggests that mean pooling, despite appearing coarse, works well in contrastive fine-tuned text encoders.

\section{How Do Fine-Tuned Text Encoders Avoid Collapse?}\label{sec:how_avoid_collapse}
In this section, we investigate how fine-tuned text encoders avoid collapse induced by mean pooling.
We argue that this robustness lies in the concentration of token embeddings within each text, as shown in the case of \GTEbase in Figure~\ref{fig:overview} (bottom).
Specifically, we theoretically show that token embeddings within each text can become concentrated under certain conditions in a simplified Transformer formulation, and that such concentration reduces \OurTool (\S~\ref{subsec:theoretical_study}).
Then, we empirically examine whether fine-tuned text encoders behave consistently with this theoretical account (\S~\ref{subsec:empirical_study}).

\subsection{Theoretical Study}\label{subsec:theoretical_study}
In the following, using a simplified Transformer formulation, we formally show how token embeddings become concentrated within each text under certain conditions, thereby lowering \OurTool.
 
To analyze how Transformer-layer components contribute to token embedding concentration within a text, we introduce a simplified formulation based on a single-head self-attention block.
This formulation abstracts the layer into three components: self-attention with projection, residual connection, and a per-token transformation.
\begin{assumption}[Setting]\label{assump:setting}
Let $\bm{H} = [\bm{h}_1, \cdots, \bm{h}_n] \in \mathbb{R}^{d \times n}$ denote the input token embeddings for a text.
The self-attention and projection components produce the attention-branch output $\bm{Z}$:
\begin{equation}
    \bm{Z} = \bm{W}^o\bm{W}^v\bm{H}\bm{A}^\top \in \mathbb{R}^{d \times n},
\end{equation}
where $\bm{A} \in \mathbb{R}^{n \times n}$ is the attention weight matrix after softmax, $d_v$ is the value dimension, and $\bm{W}^v \in \mathbb{R}^{d_v \times d}$ and $\bm{W}^o \in \mathbb{R}^{d \times d_v}$ are the value and output projection matrices, respectively.\footnote{For simplicity, we omit bias terms from $\bm{W}^v$ and $\bm{W}^o$.}
The residual connection adds $\bm{Z}$ to the input $\bm{H}$, yielding the post-residual embeddings $\bm{Y}$:
\begin{equation}
    \bm{Y} = \bm{Z} + \bm{H} \in \mathbb{R}^{d \times n}.
\end{equation}
Finally, a per-token transformation $\bm{g}$, which abstracts post-attention operations such as LayerNorm and FFN, is applied to each $\bm{y}_i$ to produce the final token embeddings $\bm{X}$:
\begin{equation}
    \bm{x}_i = \bm{g}(\bm{y}_i), \quad \bm{X} = [\bm{x}_1, \cdots, \bm{x}_n] \in \mathbb{R}^{d \times n}.
\end{equation}
We assume $\bm{h}_1, \cdots, \bm{h}_n \overset{\mathrm{i.i.d.}}{\sim} \mathcal{N}(\bm{\eta}, c\bm{I}_d) \quad (\bm{\eta} \neq \bm{0}, c>0)$ and treat $\bm{A}$ as fixed for analytical tractability.
We write $\mathbb{E}_{\bm{H}}[\cdot]$ for the expectation over the randomness of $\bm{H} = [\bm{h}_1, \ldots, \bm{h}_n]$ induced by $\bm{h}_1, \cdots, \bm{h}_n \overset{\mathrm{i.i.d.}}{\sim} \mathcal{N}(\bm{\eta}, c\bm{I}_d)$.
We also define the spread of any matrix $\bm{M} = [\bm{m}_1, \cdots, \bm{m}_n]$ to quantify the degree of token embedding concentration:
\begin{equation}
    S(\bm{M}) \coloneqq \frac{1}{n}\sum_{j=1}^{n}\|\bm{m}_j - \MU{\bm{M}}\|_2^2.
\end{equation}
\end{assumption}

Based on this formulation, token embedding concentration can arise under the following three conditions:
(i) the self-attention and projection components reduce the spread of the attention output $\bm{Z}$,
(ii) the residual connection preserves this concentration in the residual output $\bm{Y}$, and
(iii) the per-token transformation $\bm{g}$ does not excessively increase the spread of the final token embeddings $\bm{X}$.
We formalize these conditions as follows.

\begin{definition}[Spread expansion/contraction under $\bm{A}$, $\bm{W}^v$, and $\bm{W}^o$]\label{assump:attention}
    Let $\bm{1} \in \mathbb{R}^n$ be the all-ones vector and
    $\bm{P} = \bm{I}_n - \frac{1}{n}\bm{1}\bm{1}^\top \in \mathbb{R}^{n \times n}$.
    Define
    \begin{equation}
        \lambda := \|\bm{W}^o\bm{W}^v\|_\mathrm{op}^2 \frac{\|\bm{P}\bm{A}\|_F^2}{n-1},
    \end{equation}
    where $\|\cdot\|_\mathrm{op}$ denotes the operator norm and $\|\cdot\|_F$ the Frobenius norm.
\end{definition}
\begin{intuition}
    This $\lambda$ relates the spread of the attention-branch output $\bm{Z}$ to that of the input $\bm{H}$; a smaller $\lambda$ indicates that $\bm{Z}$ is more concentrated.
\end{intuition}
\begin{definition}[Ratio of token spread in input $\bm{H}$ to the scale of the residual output $\bm{Y}$]\label{assump:residual}
    Define
   \begin{equation}
    r := \frac{\mathbb{E}_{\bm{H}}\left[S(\bm{H})\right]}{\mathbb{E}_{\bm{H}}\left[\|\MU{\bm{Y}}\|_2^2\right]}.
\end{equation}
\end{definition}
\begin{intuition}
    A smaller value of $r$ indicates that the influence of the spread in the input $\bm{H}$ is smaller in the residual output $\bm{Y}=\bm{Z}+\bm{H}$.
\end{intuition}
\begin{definition}[Relative spread ratio of $\bm{X}$ to $\bm{Y}$ through transformation $\bm{g}$]\label{assump:ffn}
    Let $C > 0$ denote the smallest constant such that
    \begin{equation}
        \frac{\mathbb{E}_{\bm{H}}\left[S(\bm{X})\right]}{\mathbb{E}_{\bm{H}}\left[\|\MU{\bm{X}}\|_2^2\right]} \leq C \frac{\mathbb{E}_{\bm{H}}\left[S(\bm{Y})\right]}{\mathbb{E}_{\bm{H}}\left[\|\MU{\bm{Y}}\|_2^2\right]}.
    \end{equation}
\end{definition}
\begin{intuition}
    A smaller value of $C$ indicates that the per-token transformation $\bm{g}$ does not significantly increase the relative spread from $\bm{Y}$ to $\bm{X}$.
\end{intuition}
\begin{theorem}[Token embeddings can become concentrated within each text]\label{theo:token_concentration}
    Using Definitions~\ref{assump:attention}--\ref{assump:ffn}, if $\lambda < 1$, then the normalized spread of the final token embeddings satisfies
    \begin{equation}
        \frac{\mathbb{E}_{\bm{H}}\left[S(\bm{X})\right]}{\mathbb{E}_{\bm{H}}\left[\|\MU{\bm{X}}\|_2^2\right]} = O(rC) \quad (r, C \to 0).
    \end{equation}
\end{theorem}
\begin{intuition}
     When $\lambda < 1$, if $r$ and $C$ are also small, $\mathbb{E}_{\bm{H}}[S(\bm{X})] / \mathbb{E}_{\bm{H}}[\|\MU{\bm{X}}\|_2^2]$ becomes small, meaning that token embeddings concentrate.
\end{intuition}
\begin{proof} 
      $\bm{Z}$ is concentrated after passing through the self-attention and projection (captured by $\lambda$); the influence of the spread of $\bm{H}$ on the residual output $\bm{Y}=\bm{Z}+\bm{H}$ becomes relatively small (captured by $r$); and $\bm{g}$ does not disperse $\bm{Y}$ (captured by $C$), so $\bm{X}$ remains concentrated.
    A detailed proof is provided in Appendix~\ref{sec:proof_of_token_concentration}.
\end{proof}
 
Having established conditions for token embedding concentration, we now connect this concentration result to \OurTool.
\begin{theorem}[Token embedding concentration leads to low \OurTool]\label{theo:concentration_reduces_socm}
    Let $\bm{f}$ be a model that outputs token embeddings $\bm{X}_i = [\bm{x}_{i,1}, \cdots, \bm{x}_{i,n_i}] = \bm{f}(t_i)$ for input texts $t_1$ and $t_2$.
    Suppose that, for $i = 1, 2$, $S(\bm{X}_i)/\|\MU{\bm{X}_i}\|_2^2 < \varepsilon$.
    Then, the \OurTool value for $(\bm{X}_1^{\mathrm{norm}}, \bm{X}_2^{\mathrm{norm}})$ (Eq.~\eqref{eq:normalize_token_embedding}) satisfies $\mathrm{SOCM} = O(\varepsilon) \quad (\varepsilon \to 0)$.
\end{theorem}
\begin{intuition}
    That is, when $S(\bm{X}) / \|\MU{\bm{X}}\|_2^2$ is small, meaning that token embeddings concentrate within each text, \OurTool also becomes small.
\end{intuition}
\begin{proof}
    When $\bm{X}$ is concentrated, the token embeddings are nearly identical within each text, so $d_\Sigma$ is small, and hence \OurTool is small.
    A detailed proof is provided in Appendix~\ref{sec:proof_of_concentration_reduces_socm}.
\end{proof}
 
\subsection{Empirical Study}\label{subsec:empirical_study}
\begin{figure*}[t]
    \centering
     \begin{subfigure}[b]{0.24\textwidth}
        \centering
        \includegraphics[width=\textwidth]{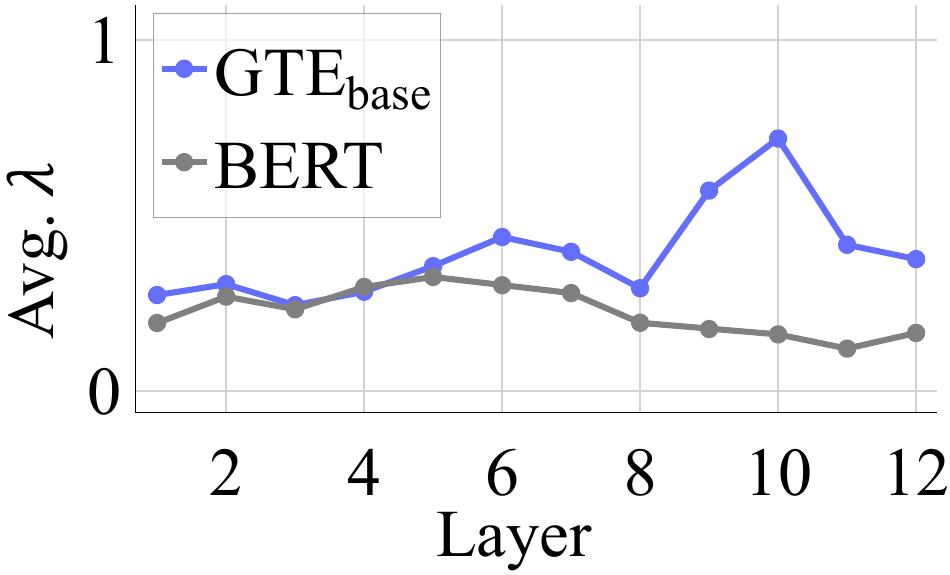}
        \caption{Avg.\ $\lambda$.}
        \label{fig:lambda_gtebase_bert}
    \end{subfigure}
    \hfill
    \begin{subfigure}[b]{0.24\textwidth}
        \centering
        \includegraphics[width=\textwidth]{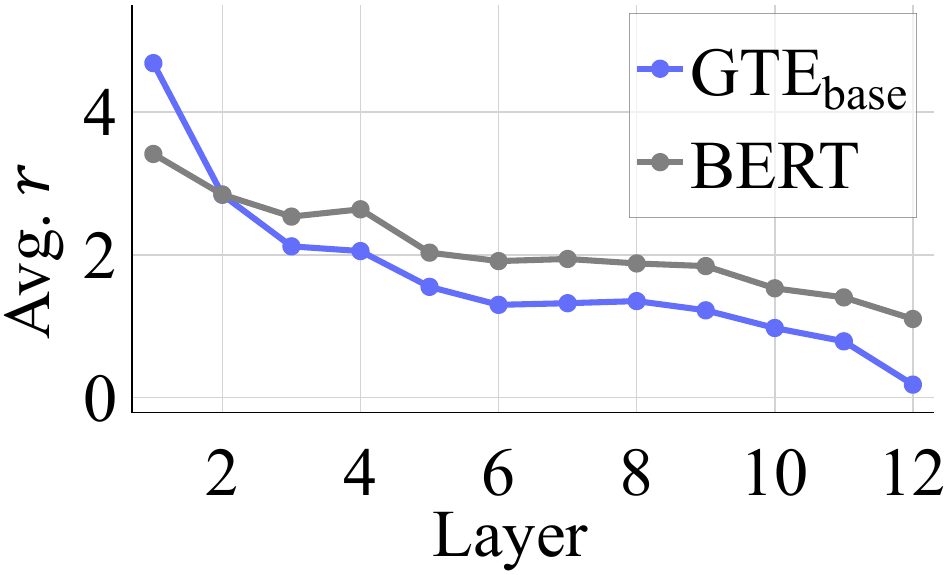}
        \caption{Avg.\ $r$.}
        \label{fig:r_gtebase_bert}
    \end{subfigure}
    \hfill
    \begin{subfigure}[b]{0.24\textwidth}
        \centering
        \includegraphics[width=\textwidth]{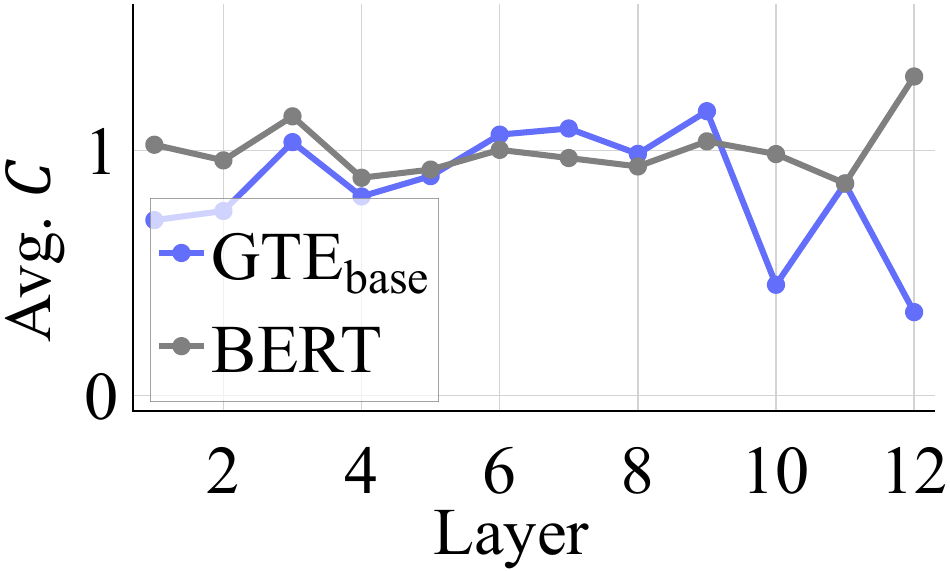}
        \caption{Avg.\ $C$.}
        \label{fig:c_gtebase_bert}
    \end{subfigure}
    \hfill
    \begin{subfigure}[b]{0.24\textwidth}
        \centering
        \includegraphics[width=\textwidth]{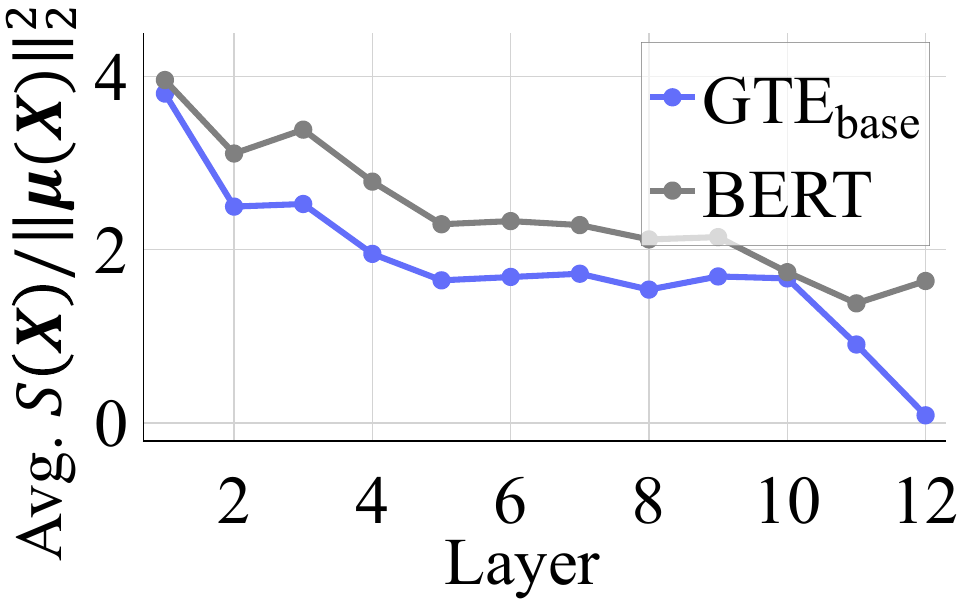}
        \caption{Avg.\ $S(\bm{X})/\|\MU{\bm{X}}\|_2^2$.}
        \label{fig:token_concentration_gtebase_bert}
    \end{subfigure}
    \caption{
    Layer-wise analysis of token embedding concentration for BERT and \GTEbase.
    (a) Avg.\ $\lambda$: a smaller value indicates greater concentration of $\bm{Z}$ relative to the input $\bm{H}$.
    (b) Avg.\ $r$: a smaller value indicates a smaller influence of the spread in $\bm{H}$ on the residual output $\bm{Y}=\bm{Z}+\bm{H}$.
    (c) Avg.\ $C$: a smaller value indicates a smaller increase in relative spread of token embeddings from $\bm{Y}$ to $\bm{X}$ through the per-token transformation $\bm{g}$.
    (d) Avg.\ $S(\bm{X})/\|\MU{\bm{X}}\|_2^2$: a smaller value indicates greater concentration of token embeddings within each text.
}
    \label{fig:how_avoid_collapse_gtebase_bert}
\end{figure*}

In the theoretical study, we showed that if the self-attention and projection term satisfies $\lambda < 1$, then the normalized spread of the final token embeddings is $O(rC)$, where $r$ and $C$ characterize the residual connection and the per-token transformation, respectively. 
Thus, when $r$ and $C$ are small, token embeddings concentrate, leading to lower \OurTool.
In the following, we empirically examine each of these quantities in actual models.

\paragraph{Setting}
We used the same 1,000 Wikipedia texts as in \S~\ref{sec:experiment}.
While \S~\ref{subsec:theoretical_study} considers a simplified setting, we computed the corresponding quantities from the actual models: $\bm{H}$ for the input hidden states to the layer, $\bm{Y}$ for the residual output combining the input hidden states with the output of the actual attention and projection operations, and $\bm{X}$ for the output hidden states, together with the attention weights $\bm{A}$ and projection matrices $\bm{W}^o$ and $\bm{W}^v$.
In the following, we compare BERT and \GTEbase (see Appendix~\ref{sec:appendix_additional_result_for_robustness} for results on other models).

\paragraph{Self-Attention and Projection ($\lambda$)}
We verify whether $\lambda < 1$ holds in practice.
Figure~\ref{fig:lambda_gtebase_bert} shows the layer-wise average of $\lambda$ for BERT and \GTEbase.
In both models, $\lambda < 1$ held across all layers.

\paragraph{Residual Connection ($r$)}
We examine the empirical counterpart of $r$, which compares the spread of the input token embeddings with the scale of the residual output.
Figure~\ref{fig:r_gtebase_bert} shows the layer-wise average of $r \coloneqq \frac{S(\bm{H})}{\|\MU{\bm{Y}}\|_2^2}$ for BERT and \GTEbase.
In \GTEbase, $r$ was lower than in BERT in some later layers and decreased toward zero in the final layer.

\paragraph{Per-Token Transformation ($C$)}
We examine the empirical counterpart of $C$, which measures the degree to which the per-token transformation $\bm{g}$ disperses token embeddings.
Figure~\ref{fig:c_gtebase_bert} shows the layer-wise average of $C \coloneqq \frac{S(\bm{X}) / \|\MU{\bm{X}}\|_2^2}{S(\bm{Y}) / \|\MU{\bm{Y}}\|_2^2}$ for BERT and \GTEbase.
$C$ was broadly similar in scale across both models, though \GTEbase showed smaller values in some later layers.

\paragraph{Token Embedding Concentration}
We examine the empirical counterpart of the concentration measure $S(\bm{X}) / \|\MU{\bm{X}}\|_2^2$.
Figure~\ref{fig:token_concentration_gtebase_bert} shows the layer-wise average of $S(\bm{X}) / \|\MU{\bm{X}}\|_2^2$ for BERT and \GTEbase.
This value was lower in \GTEbase than in BERT, especially in the later layers.
This is consistent with Theorem~\ref{theo:token_concentration}, given the observed values of $\lambda$, $r$, and $C$.
Also, the concentration in the final layer is consistent with the low \OurTool in \GTEbase, as implied by Theorem~\ref{theo:concentration_reduces_socm}.
Looking more closely at the final layer, both $r$ and $S(\bm{X}) / \|\MU{\bm{X}}\|_2^2$ were closer to zero in \GTEbase than in BERT.
Here, a smaller $r$ indicates a smaller influence of the spread in $\bm{H}$ on $\bm{Y}$.
By contrast, $\lambda$ was below 1 in both \GTEbase and BERT, suggesting that the attention and projection branch tends to concentrate token embeddings in both models.
These observations suggest that the attention and projection components may tend to concentrate token embeddings both before and after fine-tuning, while the influence of input spread in the residual connection may be smaller after fine-tuning, resulting in more concentrated output token embeddings.

\paragraph{Why Concentration Occurs in Fine-Tuned Encoders}
One possible explanation for concentration in fine-tuned encoders is that contrastive learning supervises mean-pooled embeddings.
In this setting, contrastive learning does not directly supervise individual token embeddings, but optimizes them through the resulting text embedding.
Because contrastive learning encourages this embedding to remain discriminative~\cite{Wang2020-ru}, representations are more useful when discriminative properties are reflected stably in the mean.
One way this can happen is for token embeddings within a text to become more concentrated around their mean, so that discriminative properties are more directly reflected in it.
From this perspective, in fine-tuned models, the attention component may promote concentration by aggregating information from other tokens, and this concentration may also be preserved in the residual connection.

\paragraph{Connection to Prior Work}
These results are consistent with prior work on the geometry of token embeddings in fine-tuned text encoders~\citep{Xiao2023-vu}.
\citet{Xiao2023-vu} reported that contrastive fine-tuning leads to anisotropic token embeddings within each text, particularly in the later layers.
Our findings connect this observation with robustness against collapse induced by mean pooling.

\section{Correlation with Downstream Task Performance}\label{sec:correlation_downstream_task}
In this section, we examine whether robustness to collapse by mean pooling, as quantified by \OurTool, correlates with downstream task performance.
 
\paragraph{Setting}
As a downstream task, we used MTEB (eng, v2)~\cite{Enevoldsen2025-hc}, a standard text embedding evaluation benchmark consisting of 41 tasks.
For each model used in \S~\ref{sec:experiment}, we compared its MTEB score with the \OurTool in Table~\ref{tab:collapse_wiki}.
 
\paragraph{Result}
\begin{figure}[t]  
    \centering
    \includegraphics[width=0.9\linewidth]{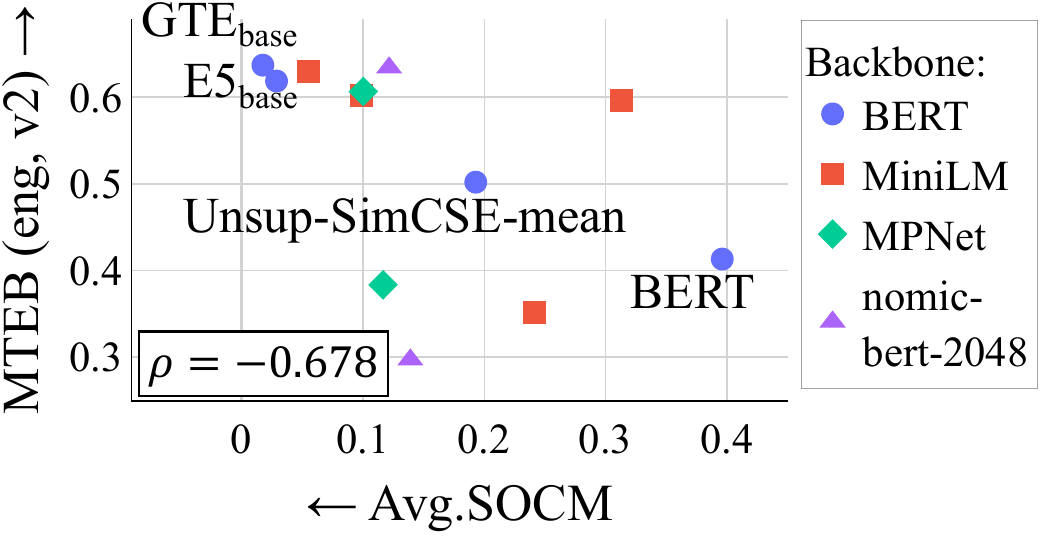}
    \caption{        
        Scatter plot of average \OurTool and MTEB (eng, v2) score.
        Each point represents a model.
        Marker types indicate the backbones.
        Models with a BERT backbone are annotated with their names.
        }
    \label{fig:corr_downstream_task_performance}
\end{figure} 

Figure~\ref{fig:corr_downstream_task_performance} shows a scatter plot of \OurTool and MTEB score for each model.
We observed a negative correlation (Spearman's $\rho = -0.678$, $p = 0.015$), indicating that models with lower \OurTool tended to achieve higher downstream task performance.
This suggests that robustness to collapse may be one factor behind the success of modern text encoders on downstream tasks.

\paragraph{Discussion}
\begin{table}[t]
    \centering
    \small
    \begin{tabular}{lc}
        \toprule
        & $\rho$ \\
        \midrule
        \OurTool & $\mathbf{-0.678}$ \\
        $S(\bm{X}) / \|\MU{\bm{X}}\|_2^2$ & $-0.622$ \\
        \bottomrule
    \end{tabular}
    \caption{
        Spearman's $\rho$ between MTEB (eng, v2) scores and each of \OurTool and $S(\bm{X})/\|\MU{\bm{X}}\|_2^2$.
    }
    \label{tab:correlation_comparison}
\end{table}

We also compared \OurTool with $S(\bm{X})/\|\MU{\bm{X}}\|_2^2$, the token concentration measure introduced in \S~\ref{sec:how_avoid_collapse}, in terms of their correlations with downstream task performance.
Table~\ref{tab:correlation_comparison} shows that \OurTool is more negatively correlated with downstream task performance.
One possible reason is that $S(\bm{X})/\|\MU{\bm{X}}\|_2^2$ does not capture inter-text separation.
As a result, this measure can be small even when semantically dissimilar texts have similar mean-pooled text embeddings.
By contrast, \OurTool also decreases as $d_\mu$ increases, thereby capturing this separation.
This suggests that, by reflecting the separation of negative pairs encouraged by contrastive fine-tuning~\cite{Wang2020-ru}, \OurTool may better capture improvements in downstream task performance brought by fine-tuning.
Further investigation into how contrastive fine-tuning improves downstream task performance remains an important direction for future work.

\section{Conclusion} \label{sec:conclusion}
This paper offered a new perspective on why modern text encoders remain effective despite adopting seemingly coarse mean pooling.
First, we argued that mean pooling can discard second-order statistics, causing distinct token embedding distributions to collapse into similar text embeddings.
We then proposed a metric to quantify this collapse (\S~\ref{sec:mean_pooling_collapse}).
Using this metric, we empirically found that contrastive fine-tuned text encoders are less prone to collapse than their backbone models in practice (\S~\ref{sec:experiment}).
We also found that the robustness of fine-tuned text encoders against this collapse lies in the concentration of token embeddings within each text (\S~\ref{sec:how_avoid_collapse}).
We further observed that this metric correlates with downstream task performance (\S~\ref{sec:correlation_downstream_task}).

\paragraph{Future Work}
A fuller mathematical account of why contrastive fine-tuning reduces \OurTool and improves downstream performance remains an important direction.
\OurTool may also be useful for encoder development; one direction is to use it as a regularization term during training.
Another direction is to extend this analysis to LLM-based generation, for example, by examining context compression by mean pooling~\cite{Feldman2025-va}.

\clearpage
\section*{Limitations}
This work has several limitations.

\paragraph{Restriction to Second-order Statistics}
This paper approximates the information in token embedding lists using only first- and second-order moments.
Specifically, we approximate token embedding lists with Gaussian distributions $\mathcal{N}(\MU{\bm{X}_i},\SIGMA{\bm{X}_i})$, which are fully characterized by their first- and second-order moments.
However, the actual distribution of token embedding lists may also contain information from third-order and higher moments.
Analyzing third-order and higher moments in token embedding lists remains an interesting direction for future work.

\paragraph{Assumptions in the Proposed Metric}
In defining the proposed metric \OurTool, we impose the assumptions $\|\MU{\bm{X}_i}\|_2=1$ and $\mathrm{tr}(\SIGMA{\bm{X}_i})\leq 2$.
As shown in Appendix~\ref{sec:appendix_trace_bound}, these assumptions hold under the normalization procedure in \S~\ref{sec:experiment} and certain conditions on model architecture (LayerNorm with shared parameters across dimensions and sufficient similarity of token embeddings within texts).
However, the proposed metric may not work effectively in cases where these assumptions are violated, for instance, when the second-order moment $\mathrm{tr}(\SIGMA{\bm{X}_i})$ becomes extremely large.
Extending the generality of the metric to broader settings remains future work.

\paragraph{Toward Theoretical Understanding}
In \S~\ref{sec:how_avoid_collapse}, we partially addressed the theoretical understanding of the observed robustness.
Specifically, we showed theoretically that token embedding concentration leads to lower \OurTool, and empirically verified that fine-tuned text encoders satisfy the conditions under which such concentration occurs.
However, why contrastive fine-tuning induces token embedding concentration in the first place remains an open question.
A deeper theoretical investigation of this mechanism is an interesting direction for future work.

\paragraph{Practical Utility of SOCM}
In this paper, we use SOCM as an analysis tool for characterizing robustness to collapse by mean pooling across models and texts.
An interesting direction for future work is to explore practical uses of SOCM beyond analysis, such as incorporating it into training objectives as a regularization term.

\paragraph{Scope Limited to Text Embeddings}
This paper focuses on text embeddings and does not address LLM-based applications, such as generation or reasoning.
Broadening the scope to these LLM-based applications, for example, by investigating how context compression by mean pooling~\cite{Feldman2025-va} influences generation or reasoning, remains an interesting direction for future work.

\paragraph{Focus on Mean Pooling}
This paper focuses on mean pooling as the target aggregation method.
Mean pooling remains the dominant pooling strategy in modern text encoders.
However, richer pooling methods, such as SIF weighting~\cite{Arora2017-pn}, have also been proposed as alternatives to simple averaging.
Investigating such methods is an interesting direction for future work.

\section*{Ethical Consideration}
This study analyzes embeddings using models released under the MIT License (MiniLM, $\text{E5}_\text{small}$, $\text{E5}_\text{base}$, $\text{GTE}_\text{small}$, $\text{GTE}_\text{base}$, Unsupervised SimCSE, MPNet) and Apache License 2.0 (all-MiniLM-L12-v2, all-mpnet-base-v2, nomic-bert-2048, nomic-embed-text-v1.5).
Our analysis, which examines the embedding representations output by these models for given input texts, falls within their intended use cases.
We use datasets (Wikipedia, MS MARCO, MTEB (eng, v2)) released under Apache License 2.0.
These datasets are used as provided without additional preprocessing to remove social biases, personal information, or offensive content, and thus may reflect various biases present in the original data.

\section*{Acknowledgments}
This work was supported by AMED Grant Number JP26wm0625405, JSPS KAKENHI Grant Number JP22H05106, the JST FOREST Program Grant Number JPMJFR2331, and JST BOOST Grant Number JPMJBS2421.
We would like to thank the members of the Tohoku NLP Group, as well as the organizers and participants of IBIS 2025 and NLP 2026, for their insightful and encouraging feedback.

\bibliography{reference}

\clearpage
\appendix
\section{Gaussian Characterization}
\label{sec:gaussian_characterization}
As noted in \S~\ref{sec:mean_pooling_collapse}, our method evaluates the target distributions by approximating them with Gaussian distributions.
This is an intentional design choice motivated by computational tractability and stability in Wasserstein-based evaluation.

\paragraph{Advantages of the Gaussian Characterization}
First, when using the Wasserstein distance, the Gaussian characterization offers substantial advantages despite ignoring higher-order moments.
For general non-Gaussian distributions, computing the Wasserstein distance typically requires costly linear programming-based algorithms whose complexity scales poorly with dimensionality, making them impractical in high dimensions~\cite{Cuturi2013-bj}.
From this perspective, the Gaussian characterization is essential for efficient computation in high-dimensional spaces, and its computational and analytical benefits outweigh the limitation of not explicitly modeling higher-order moments.

\paragraph{Information Captured by First- and Second-order Statistics}
Furthermore, focusing on first- and second-order statistics lets us efficiently capture much of the information in the data distribution.
In particular, it is known that the $L_2$-Wasserstein distance computed using only up to second-order statistics can provide a universal lower bound on the true $L_2$-Wasserstein distance between arbitrary distributions~\cite{Cuesta-Albertos1996-bd}.
In this sense, introducing a Gaussian characterization is not merely a heuristic simplification; rather, it recovers information that is theoretically guaranteed in terms of the Wasserstein distance up to second-order statistics.
Moreover, such characterizations are widely used in practice.
For example, the Fr\'echet Inception Distance (FID), a standard metric for evaluating generative models, also relies on the $L_2$-Wasserstein distance between Gaussian approximations characterized by first- and second-order statistics, and has been successfully used across a broad range of distributions~\cite{Heusel2017-ic}.

Taken together, these theoretical guarantees and empirical precedents support the use of second-order statistics-based Wasserstein characterization as a principled and practically effective approach.

\section{Proof that \OurTool Satisfies Desirable Properties} \label{sec:appendix_proof}
In this section, we prove that \OurTool, as defined in Eq.~\eqref{eq:socm}, satisfies the desirable properties (a)–(e) introduced in \S~\ref{subsubsec:properties}.
Specifically, we show that
\begin{equation}
    \mathrm{\OurTool}(d_\mu, d_\Sigma) = (1 - d_\mu) d_\Sigma
\end{equation}
satisfies all properties (a)–(e):
\begin{itemize}
    \item[(a)] $\displaystyle d_\mu = 0 \land d_\Sigma = 1 \Leftrightarrow \mathrm{\OurTool} = 1$
    \item[(b)] $\displaystyle d_\mu = 1 \lor d_\Sigma = 0 \Leftrightarrow \mathrm{\OurTool} = 0$
    \item[(c)] $\displaystyle \frac{\partial\,\mathrm{\OurTool}}{\partial d_\mu} \leq 0$
    \item[(d)] $\displaystyle \frac{\partial\,\mathrm{\OurTool}}{\partial d_\Sigma} \geq 0$
    \item[(e)] $\displaystyle \frac{\partial^2\,\mathrm{\OurTool}}{\partial d_\mu\,\partial d_\Sigma} \leq 0$
\end{itemize}

\subsection{Proof of Property (a)}
We prove property (a): $d_\mu = 0 \land d_\Sigma = 1 \Leftrightarrow \mathrm{\OurTool} = 1$.
We first show $(d_\mu = 0 \land d_\Sigma = 1 \Rightarrow \mathrm{\OurTool} = 1)$.
When $d_\mu = 0$ and $d_\Sigma = 1$,
\begin{align}
    \mathrm{\OurTool}(0, 1) 
    &= (1 - 0) \cdot 1 \\
    &= 1.
\end{align}
Therefore, $d_\mu = 0 \land d_\Sigma = 1 \Rightarrow \mathrm{\OurTool} = 1$ holds.
Next, we show $(\mathrm{\OurTool} = 1 \Rightarrow d_\mu = 0 \land d_\Sigma = 1)$.
Assume $\mathrm{\OurTool}(d_\mu, d_\Sigma) = 1$, i.e.,
\begin{equation}
    (1 - d_\mu) d_\Sigma = 1.
\end{equation}
Since $d_\mu, d_\Sigma \in [0, 1]$, we have $(1 - d_\mu) \in [0, 1]$ and $d_\Sigma \in [0, 1]$.
Therefore, the above equation holds only when $1 - d_\mu = 1$ and $d_\Sigma = 1$, which implies $d_\mu = 0$ and $d_\Sigma = 1$.
Therefore, $\mathrm{\OurTool} = 1 \Rightarrow d_\mu = 0 \land d_\Sigma = 1$ holds.

\subsection{Proof of Property (b)}
We prove property (b): $d_\mu = 1 \lor d_\Sigma = 0 \Leftrightarrow \mathrm{\OurTool} = 0$.
We first show $(d_\mu = 1 \lor d_\Sigma = 0 \Rightarrow \mathrm{\OurTool} = 0)$.
When $d_\mu = 1$,
\begin{align}
    \mathrm{\OurTool}(1, d_\Sigma) 
    &= (1 - 1) \cdot d_\Sigma \\
    &= 0.
\end{align}
Also, when $d_\Sigma = 0$,
\begin{align}
    \mathrm{\OurTool}(d_\mu, 0) 
    &= (1 - d_\mu) \cdot 0 \\
    &= 0.
\end{align}
Therefore, $d_\mu = 1 \lor d_\Sigma = 0 \Rightarrow \mathrm{\OurTool} = 0$ holds.
Next, we show $(\mathrm{\OurTool} = 0 \Rightarrow d_\mu = 1 \lor d_\Sigma = 0)$.
Assume $\mathrm{\OurTool}(d_\mu, d_\Sigma) = 0$, i.e.,
\begin{equation}
    (1 - d_\mu) d_\Sigma = 0.
\end{equation}
This holds when $1 - d_\mu = 0$ or $d_\Sigma = 0$, i.e., when $d_\mu = 1$ or $d_\Sigma = 0$.
Therefore, $\mathrm{\OurTool} = 0 \Rightarrow d_\mu = 1 \lor d_\Sigma = 0$ holds.

\subsection{Proof of Property (c)}
We prove property (c): $\displaystyle \frac{\partial\,\mathrm{\OurTool}}{\partial d_\mu} \leq 0$.
Taking the partial derivative of $\mathrm{\OurTool}(d_\mu, d_\Sigma) = (1 - d_\mu) d_\Sigma$ with respect to $d_\mu$,
\begin{equation}
    \frac{\partial\,\mathrm{\OurTool}}{\partial d_\mu} = -d_\Sigma.
\end{equation}
Since $d_\Sigma \in [0, 1]$, we have $-d_\Sigma \leq 0$.
Therefore, $\displaystyle \frac{\partial\,\mathrm{\OurTool}}{\partial d_\mu} \leq 0$ holds.

\subsection{Proof of Property (d)}
We prove property (d): $\displaystyle \frac{\partial\,\mathrm{\OurTool}}{\partial d_\Sigma} \geq 0$.
Taking the partial derivative of $\mathrm{\OurTool}(d_\mu, d_\Sigma) = (1 - d_\mu) d_\Sigma$ with respect to $d_\Sigma$,
\begin{equation}
    \frac{\partial\,\mathrm{\OurTool}}{\partial d_\Sigma} = 1 - d_\mu.
\end{equation}
Since $d_\mu \in [0, 1]$, we have $1 - d_\mu \geq 0$.
Therefore, $\displaystyle \frac{\partial\,\mathrm{\OurTool}}{\partial d_\Sigma} \geq 0$ holds.

\subsection{Proof of Property (e)}
We prove property (e): $\displaystyle \frac{\partial^2\,\mathrm{\OurTool}}{\partial d_\mu\,\partial d_\Sigma} \leq 0$.
From the proof of property (c),
\begin{equation}
    \frac{\partial\,\mathrm{\OurTool}}{\partial d_\mu} = -d_\Sigma.
\end{equation}
Taking the partial derivative with respect to $d_\Sigma$,
\begin{equation}
    \frac{\partial^2\,\mathrm{\OurTool}}{\partial d_\mu\,\partial d_\Sigma} = -1 \leq 0.
\end{equation}
Therefore, property (e) holds.

\subsection{Generalization}
More generally, for any $\alpha > 0$ and $\beta > 0$, the following form also satisfies all properties (a)–(e):
\begin{equation}
    \mathrm{\OurTool}_{\alpha,\beta}(d_\mu, d_\Sigma) = (1 - d_\mu)^{\alpha} d_\Sigma^{\beta}.
\end{equation}
The proof follows analogously, replacing $1 - d_\mu$ and $d_\Sigma$ with their respective powers.
Our adopted form in Eq.~\eqref{eq:socm} corresponds to the simple case $\alpha = \beta = 1$.

\section{Proof of Normalization Property} \label{sec:appendix_normalization}
We prove that under the normalization defined in \S~\ref{sec:experiment}, the relationship $\MU{\bm{X}_{i}^\text{norm}} = \MU{\bm{X}_{i}}/\|\MU{\bm{X}_{i}}\|_2$ holds.

\paragraph{Statement}
Given a token embedding list $\bm{X}_{i} = [\bm{x}_{i,1}, \ldots, \bm{x}_{i,n_{i}}] \in \mathbb{R}^{d \times n_{i}}$ and its normalized version
\begin{equation}
    \bm{X}_{i}^\text{norm} = \left[\frac{\bm{x}_{i,1}}{\|\MU{\bm{X}_{i}}\|_2}, \ldots, \frac{\bm{x}_{i,n_{i}}}{\|\MU{\bm{X}_{i}}\|_2}\right] \in \mathbb{R}^{d \times n_{i}},
\end{equation}
we show that $\MU{\bm{X}_{i}^\text{norm}} = \MU{\bm{X}_{i}}/\|\MU{\bm{X}_{i}}\|_2$.

\paragraph{Proof}
By the definition of mean pooling, we have
\begin{align}
    \MU{\bm{X}_{i}^\text{norm}} 
    &= \frac{1}{n_i}\sum_{j=1}^{n_i} \frac{\bm{x}_{i,j}}{\|\MU{\bm{X}_{i}}\|_2}.
\end{align}
Since $\|\MU{\bm{X}_{i}}\|_2$ is a scalar independent of the summation index $j$, we can factor it out:
\begin{align}
    \MU{\bm{X}_{i}^\text{norm}} 
    &= \frac{1}{\|\MU{\bm{X}_{i}}\|_2} \cdot \frac{1}{n_i}\sum_{j=1}^{n_i} \bm{x}_{i,j}.
\end{align}
The remaining summation is precisely the definition of $\MU{\bm{X}_{i}}$:
\begin{align}
    \MU{\bm{X}_{i}^\text{norm}} 
    &= \frac{1}{\|\MU{\bm{X}_{i}}\|_2} \cdot \MU{\bm{X}_{i}} \\
    &= \frac{\MU{\bm{X}_{i}}}{\|\MU{\bm{X}_{i}}\|_2}.
\end{align}
Furthermore, this implies that $\|\MU{\bm{X}_{i}^\text{norm}}\|_2 = 1$, which is the desired property for computing \OurTool as defined in \S~\ref{subsubsec:socm_definition}.

\section{Implementation Details} \label{sec:appendix_implementation_details}
This section provides detailed information about the implementation of our experiments described in \S~\ref{sec:experiment}, \S~\ref{sec:how_avoid_collapse}, and \S~\ref{sec:correlation_downstream_task}.
\subsection{Dataset Details}

\paragraph{Preprocessing}
For comparison across different models and datasets, we did not use any task-specific prefixes (e.g., \texttt{query:}, \texttt{passage:}) when encoding texts.
Note that some text embedders are designed to utilize such prefixes to distinguish between different text types~\cite{Wang2022-gw, Li2023-oy}.

\paragraph{Dataset URLs}
Table~\ref{tab:dataset_urls} shows the URLs for the datasets used in our experiments.
\begin{table*}[t]
    \small
    \centering
    \begin{tabular}{@{}ll@{}}
        \toprule
        \textbf{Dataset} & \textbf{URL} \\ 
        \midrule
        Wikipedia & \parbox{10cm}{\url{https://huggingface.co/datasets/princeton-nlp/datasets-for-simcse/resolve/main/wiki1m_for_simcse.txt}} \\
        MS MARCO & \url{https://huggingface.co/datasets/mteb/msmarco} \\
        MS MARCO Hard Negatives & \parbox{10cm}{\url{https://huggingface.co/datasets/sentence-transformers/msmarco-co-condenser-margin-mse-sym-mnrl-mean-v1}} \\
        MTEB (eng, v2) & \url{https://github.com/embeddings-benchmark/mteb} \\
        \bottomrule
    \end{tabular}
    \caption{URLs for datasets used in our experiments.}
    \label{tab:dataset_urls}
\end{table*}

\paragraph{Language}
In our experiments, we used English-language datasets.

\subsection{Model Details}
\begin{table*}[t]
    \small
    \centering
    \begin{tabular}{l l r}
        \toprule
        \textbf{Model} & \texttt{model\_name} & \textbf{Params (M)} \\
        \midrule
        BERT & \texttt{bert-base-uncased} & 110 \\
        $\text{Unsup-SimCSE-mean}$ & \texttt{h-tomo/unsup-simcse-bert-base-uncased-mean} & 110 \\
        $\text{E5}_{\text{base}}$ & \texttt{intfloat/e5-base-v2} & 110 \\
        $\text{GTE}_{\text{base}}$ & \texttt{thenlper/gte-base} & 110 \\
        \midrule
        MiniLM & \texttt{microsoft/MiniLM-L12-H384-uncased} & 30 \\
        all-MiniLM-L12-v2 & \texttt{sentence-transformers/all-MiniLM-L12-v2} & 30 \\
        $\text{E5}_{\text{small}}$ & \texttt{intfloat/e5-small-v2} & 30 \\
        $\text{GTE}_{\text{small}}$ & \texttt{thenlper/gte-small} & 30 \\
        \midrule
        MPNet & \texttt{microsoft/mpnet-base} & 109 \\
        all-mpnet-base-v2 & \texttt{sentence-transformers/all-mpnet-base-v2} & 109 \\
        \midrule
        nomic-bert-2048 & \texttt{nomic-ai/nomic-bert-2048} & 137 \\
        nomic-embed-text-v1.5 & \texttt{nomic-ai/nomic-embed-text-v1.5} & 137 \\
        \bottomrule
    \end{tabular}
    \caption{Details of the Hugging Face models used in our experiments, with parameter counts (in millions) taken from the corresponding backbone model sizes.}
    \label{tab:model_details}
\end{table*}

Table~\ref{tab:model_details} lists the specific model identifiers from Hugging Face for each model used in our experiments.
For most models, we used publicly available pre-trained checkpoints.
For Unsupervised SimCSE~\cite{Gao2021-ds}, we trained the model ourselves using the original codebase, as the unsupervised checkpoint was not publicly available at the time of our experiments.
We followed the default training configuration provided in the official implementation, using bert-base-uncased as the backbone model and training on English Wikipedia with the default hyperparameters.

\section{Additional Results for \S~\ref{sec:experiment}}\label{sec:appendix_msmarco_results}
To validate the generalizability of our findings, we conducted additional experiments on MS MARCO.

\paragraph{Experimental Setting}
We followed the same experimental procedure as described in \S~\ref{sec:experiment}.
Specifically, we randomly sampled 1,000 texts from MS MARCO passages and generated 499,500 text pairs by comparing these texts pairwise.
We computed \OurTool values for the same set of models used in \S~\ref{sec:experiment}.
In addition, we computed \OurTool for 50,000 query-negative passage pairs from MS MARCO hard negatives.
\paragraph{Results}
\begin{table}[t]\centering
    \small
    \begin{tabular}{llr}
        \toprule
        \textbf{Model} &\textbf{Avg. \OurTool} $\downarrow$ \\
        \midrule
        $\text{BERT}$ &$0.491$ \\
        $\rightarrow \text{Unsup-SimCSE-mean}$ &$\bm{0.269}$ $(-0.222)$ \\
        $\rightarrow \text{E5}_\text{base}$ &$\bm{0.043}$ $(-0.448)$ \\
        $\rightarrow \text{GTE}_\text{base}$ &$\bm{0.025}$ $(-0.466)$ \\
        \midrule
        $\text{MiniLM}$ &$0.289$  \\
        $\rightarrow \text{all-MiniLM-L12-v2}$ &$0.348$ $\hspace{1.2mm}(+0.059)$ \\
        $\rightarrow \text{E5}_\text{small}$ &$\bm{0.085}$ $(-0.204)$ \\
        $\rightarrow \text{GTE}_\text{small}$ &$\bm{0.055}$ $(-0.234)$ \\
        \midrule
        $\text{MPNet}$ &$0.106$ \\
        $\rightarrow \text{all-mpnet-base-v2}$ &$\bm{0.094}$ $(-0.012)$ \\
        \midrule
        $\text{nomic-bert-2048}$ &$0.110$ \\
        $ \rightarrow \text{nomic-embed-text-v1.5}$ &$0.133$ $\hspace{1.2mm}(+0.023)$ \\
        \bottomrule
    \end{tabular}
    \caption{
        Average \OurTool values for each model on text pairs from MS MARCO passages.
        For text encoders derived from backbone models, values in parentheses show the change in \OurTool.
        Bold values indicate a reduction.
    }
\label{tab:collapse_msmarco}
\end{table}

\begin{table}[t]\centering
    \small
    \begin{tabular}{llr}
        \toprule
        \textbf{Model} &\textbf{Avg. \OurTool} $\downarrow$ \\
        \midrule
        $\text{BERT}$ &$0.480$ \\
        $\rightarrow \text{Unsup-SimCSE-mean}$ &$\bm{0.257}$ $(-0.224)$ \\
        $\rightarrow \text{E5}_\text{base}$ &$\bm{0.036}$ $(-0.445)$ \\
        $\rightarrow \text{GTE}_\text{base}$ &$\bm{0.017}$ $(-0.464)$ \\
        \midrule
        $\text{MiniLM}$ &$0.340$  \\
        $\rightarrow \text{all-MiniLM-L12-v2}$ &$\bm{0.330}$ $(-0.010)$ \\
        $\rightarrow \text{E5}_\text{small}$ &$\bm{0.087}$ $(-0.253)$ \\
        $\rightarrow \text{GTE}_\text{small}$ &$\bm{0.048}$ $(-0.292)$ \\
        \midrule
        $\text{MPNet}$ &$0.129$ \\
        $\rightarrow \text{all-mpnet-base-v2}$ &$\bm{0.093}$ $(-0.036)$ \\
        \midrule
        $\text{nomic-bert-2048}$ &$0.131$ \\
        $ \rightarrow \text{nomic-embed-text-v1.5}$ &$\bm{0.104}$ $(-0.027)$ \\
        \bottomrule
    \end{tabular}
    \caption{
        Average \OurTool values for each model on query--negative passage pairs from MS MARCO hard negatives.
        For text encoders derived from backbone models, values in parentheses show the change in \OurTool.
        Bold values indicate a reduction.
    }
\label{tab:collapse_hard_negative}
\end{table}

Table~\ref{tab:collapse_msmarco} shows the average \OurTool values for each model on MS MARCO, and Table~\ref{tab:collapse_hard_negative} shows those on MS MARCO hard negatives.
In both cases, we observed similar trends to the Wikipedia results (\S~\ref{sec:experiment}).

\section{Proof of Theorem~\ref{theo:token_concentration}}\label{sec:proof_of_token_concentration}
Recall that for any matrix $\bm{M} \in \mathbb{R}^{d \times n}$,
\begin{equation}
    S(\bm{M}) = \frac{1}{n}\sum_{j=1}^{n} \|\bm{m}_j - \MU{\bm{M}}\|_2^2 = \frac{1}{n}\|\bm{M}\bm{P}\|_F^2,
\end{equation}
where $\bm{P} = \bm{I}_n - \frac{1}{n}\bm{1}\bm{1}^\top$ is the centering matrix.
 
We first control the spread of $\bm{Z} = \bm{W}^o\bm{W}^v\bm{H}\bm{A}^\top$.
Since $\bm{P}$ is symmetric,
\begin{equation}
    \bm{Z}\bm{P} = \bm{W}^o\bm{W}^v\bm{H}\bm{A}^\top\bm{P} = \bm{W}^o\bm{W}^v\bm{H}(\bm{P}\bm{A})^\top.
\end{equation}
Hence, by the operator norm inequality $\|\bm{AB}\|_F \leq \|\bm{A}\|_{\mathrm{op}}\|\bm{B}\|_F$,
\begin{align}
    \mathbb{E}_{\bm{H}}[S(\bm{Z})]
    &= \frac{1}{n}\mathbb{E}_{\bm{H}}\bigl[\|\bm{Z}\bm{P}\|_F^2\bigr] \notag \\
    &= \frac{1}{n}\mathbb{E}_{\bm{H}}\bigl[\|\bm{W}^o\bm{W}^v\bm{H}(\bm{P}\bm{A})^\top\|_F^2\bigr] \notag \\
    &\leq \frac{\|\bm{W}^o\bm{W}^v\|_{\mathrm{op}}^2}{n}\mathbb{E}_{\bm{H}}\bigl[\|\bm{H}(\bm{P}\bm{A})^\top\|_F^2\bigr].
\end{align}
Write $\bm{H} = \bm{\eta}\bm{1}^\top + \bm{E}$, where the columns of $\bm{E}$ are i.i.d.\ $\mathcal{N}(\bm{0}, c\bm{I}_d)$.
Since each row of $\bm{A}$ sums to one (i.e., $\bm{A}\bm{1} = \bm{1}$), we have $(\bm{P}\bm{A})\bm{1} = \bm{P}(\bm{A}\bm{1}) = \bm{P}\bm{1} = \bm{0}$, and thus $\bm{1}^\top(\bm{P}\bm{A})^\top = \bm{0}^\top$.
Therefore, the mean part vanishes:
\begin{equation}
    \bm{H}(\bm{P}\bm{A})^\top = \bm{E}(\bm{P}\bm{A})^\top.
\end{equation}
By the isotropy of the Gaussian noise, $\mathbb{E}_{\bm{H}}\bigl[\|\bm{E}\bm{B}\|_F^2\bigr] = cd\,\|\bm{B}\|_F^2$ for any $\bm{B} \in \mathbb{R}^{n \times n}$.
Taking $\bm{B} = (\bm{P}\bm{A})^\top$,
\begin{equation}
    \mathbb{E}_{\bm{H}}[S(\bm{Z})] \leq \frac{\|\bm{W}^o\bm{W}^v\|_{\mathrm{op}}^2}{n} \cdot cd\,\|\bm{P}\bm{A}\|_F^2.
\end{equation}
On the other hand, since $\|\bm{P}\|_F^2 = \mathrm{tr}(\bm{P}) = n - 1$,
\begin{align}
    \mathbb{E}_{\bm{H}}[S(\bm{H})]
    &= \frac{1}{n}\mathbb{E}_{\bm{H}}[\|\bm{E}\bm{P}\|_F^2] \notag \\
    &= \frac{cd}{n}\|\bm{P}\|_F^2 \notag \\
    &= \frac{cd(n-1)}{n}.
\end{align}
Combining the above two inequalities,
\begin{align}
    \mathbb{E}_{\bm{H}}[S(\bm{Z})]
    &\leq \|\bm{W}^o\bm{W}^v\|_{\mathrm{op}}^2 \frac{\|\bm{P}\bm{A}\|_F^2}{n-1}\,\mathbb{E}_{\bm{H}}[S(\bm{H})] \notag \\
    &= \lambda\,\mathbb{E}_{\bm{H}}[S(\bm{H})].
\end{align}
By Definition~\ref{assump:attention} and the hypothesis $\lambda < 1$,
 
Next, we control the spread of $\bm{Y} = \bm{Z} + \bm{H}$.
Using $\bm{Y}\bm{P} = \bm{Z}\bm{P} + \bm{H}\bm{P}$ and the Minkowski inequality,
\begin{align}
    \sqrt{\mathbb{E}_{\bm{H}}[S(\bm{Y})]}
    &= \frac{1}{\sqrt{n}}\Bigl(\mathbb{E}_{\bm{H}}\bigl[\|\bm{Y}\bm{P}\|_F^2\bigr]\Bigr)^{1/2} \notag \\
    &\leq \frac{1}{\sqrt{n}}\Bigl(\mathbb{E}_{\bm{H}}\bigl[\|\bm{Z}\bm{P}\|_F^2\bigr]\Bigr)^{1/2} \notag \\
    &\quad + \frac{1}{\sqrt{n}}\Bigl(\mathbb{E}_{\bm{H}}\bigl[\|\bm{H}\bm{P}\|_F^2\bigr]\Bigr)^{1/2} \notag \\
    &= \sqrt{\mathbb{E}_{\bm{H}}[S(\bm{Z})]} + \sqrt{\mathbb{E}_{\bm{H}}[S(\bm{H})]} \notag \\
    &\leq (1 + \sqrt{\lambda})\sqrt{\mathbb{E}_{\bm{H}}[S(\bm{H})]}.
\end{align}
Squaring both sides yields
\begin{equation}
    \mathbb{E}_{\bm{H}}[S(\bm{Y})] \leq (1 + \sqrt{\lambda})^2\,\mathbb{E}_{\bm{H}}[S(\bm{H})].
\end{equation}
Therefore,
\begin{equation}
    \frac{\mathbb{E}_{\bm{H}}[S(\bm{Y})]}{\mathbb{E}_{\bm{H}}[\|\MU{\bm{Y}}\|_2^2]}
    \leq (1 + \sqrt{\lambda})^2 \frac{\mathbb{E}_{\bm{H}}[S(\bm{H})]}{\mathbb{E}_{\bm{H}}[\|\MU{\bm{Z}+\bm{H}}\|_2^2]}.
\end{equation}
By Definition~\ref{assump:residual}, it holds that $r=\frac{\mathbb{E}_{\bm{H}}[S(\bm{H})]}{\mathbb{E}_{\bm{H}}[\|\MU{\bm{Z}+\bm{H}}\|_2^2]}$, hence
\begin{equation}
    \frac{\mathbb{E}_{\bm{H}}[S(\bm{Y})]}{\mathbb{E}_{\bm{H}}[\|\MU{\bm{Y}}\|_2^2]} \leq (1 + \sqrt{\lambda})^2 r = O(r),
\end{equation}
where the hidden constant depends only on $\lambda$.
 
Finally, Definition~\ref{assump:ffn} gives
\begin{align}
    \frac{\mathbb{E}_{\bm{H}}[S(\bm{X})]}{\mathbb{E}_{\bm{H}}[\|\MU{\bm{X}}\|_2^2]}
    &\leq C\,\frac{\mathbb{E}_{\bm{H}}[S(\bm{Y})]}{\mathbb{E}_{\bm{H}}[\|\MU{\bm{Y}}\|_2^2]} \notag \\
    &\leq C(1 + \sqrt{\lambda})^2 r.
\end{align}
We therefore conclude that
\begin{equation}
    \frac{\mathbb{E}_{\bm{H}}[S(\bm{X})]}{\mathbb{E}_{\bm{H}}[\|\MU{\bm{X}}\|_2^2]} = O(rC) \quad (r, C \to 0).
\end{equation}
This completes the proof.

\section{Proof of Theorem~\ref{theo:concentration_reduces_socm}}\label{sec:proof_of_concentration_reduces_socm}
Let $t_1, t_2$ be arbitrary texts, and let $\bm{X}_i = \bm{f}(t_i)$ for $i = 1, 2$.
By definition of $\MU{\cdot}$,
\begin{equation}
    \MU{\bm{X}_i^\text{norm}} = \frac{\MU{\bm{X}_i}}{\|\MU{\bm{X}_i}\|_2},
\end{equation}
hence $\|\MU{\bm{X}_i^\text{norm}}\|_2 = 1$.
Since normalization scales each token embedding by the scalar $1/\|\MU{\bm{X}_i}\|_2$, the covariance matrix scales quadratically:
\begin{equation}
    \SIGMA{\bm{X}_i^\text{norm}} = \frac{1}{\|\MU{\bm{X}_i}\|_2^2}\SIGMA{\bm{X}_i}.
\end{equation}
Using $S(\bm{X}) = \mathrm{tr}(\SIGMA{\bm{X}})$, the assumption gives
\begin{equation}
    \mathrm{tr}\!\left(\SIGMA{\bm{X}_i^\text{norm}}\right)
    = \frac{S(\bm{X}_i)}{\|\MU{\bm{X}_i}\|_2^2}
    < \varepsilon
    \qquad (i = 1, 2).
\end{equation}
Since $(1 - d_\mu) \leq 1$, we have
\begin{equation}
    \mathrm{\OurTool} \leq d_\Sigma.
\end{equation}
By definition of $d_\Sigma$,
\begin{align}
    d_\Sigma
    &= \frac{1}{4}\mathrm{tr}\Bigl(
        \SIGMA{\bm{X}_1^\text{norm}} + \SIGMA{\bm{X}_2^\text{norm}} \notag \\
    &\quad - 2\Bigl(\SIGMA{\bm{X}_1^\text{norm}}^{1/2}
        \SIGMA{\bm{X}_2^\text{norm}} \notag \\
    &\qquad\qquad \SIGMA{\bm{X}_1^\text{norm}}^{1/2}\Bigr)^{1/2}
    \Bigr).
\end{align}
Since $\bigl(\SIGMA{\bm{X}_1^\text{norm}}^{1/2}\SIGMA{\bm{X}_2^\text{norm}}\SIGMA{\bm{X}_1^\text{norm}}^{1/2}\bigr)^{1/2}$ is positive semidefinite, its trace is nonnegative, and therefore
\begin{equation}
    d_\Sigma
    \leq \frac{1}{4}\mathrm{tr}\!\left(\SIGMA{\bm{X}_1^\text{norm}}\right)
    + \frac{1}{4}\mathrm{tr}\!\left(\SIGMA{\bm{X}_2^\text{norm}}\right)
    < \frac{\varepsilon}{2}.
\end{equation}
Hence,
\begin{equation}
    \mathrm{\OurTool}
    \leq d_\Sigma
    < \frac{\varepsilon}{2},
\end{equation}
which gives $\mathrm{\OurTool} = O(\varepsilon)$ as $\varepsilon \to 0$.
This completes the proof.

\section{Additional Results for \S~\ref{sec:how_avoid_collapse}}\label{sec:appendix_additional_result_for_robustness}
\subsection{Results for Additional Models}
\begin{figure*}[t]
    \centering
     \begin{subfigure}[b]{0.24\textwidth}
        \centering
        \includegraphics[width=\textwidth]{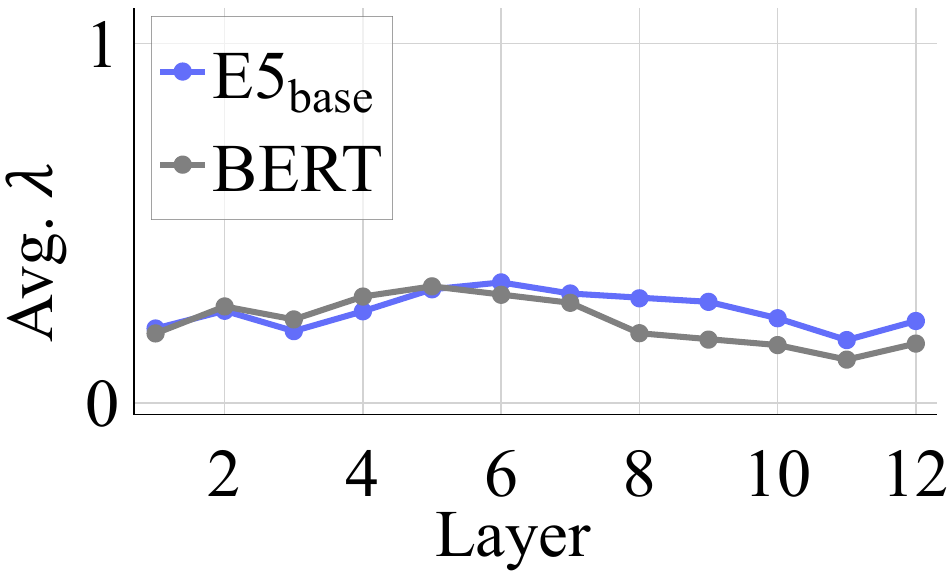}
        \caption{Avg.\ $\lambda$.}
        \label{fig:lambda_e5base_bert}
    \end{subfigure}
    \hfill
    \begin{subfigure}[b]{0.24\textwidth}
        \centering
        \includegraphics[width=\textwidth]{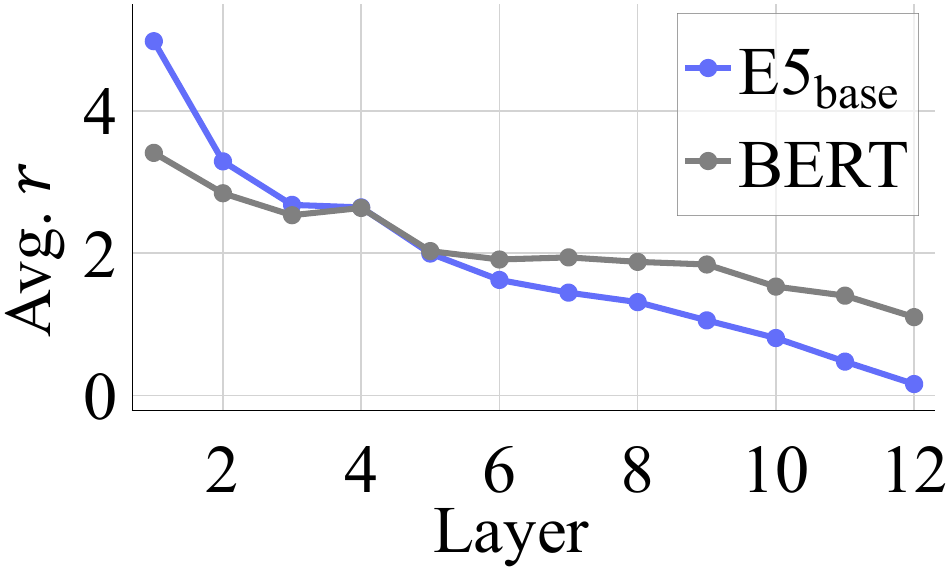}
        \caption{Avg.\ $r$.}
        \label{fig:r_e5base_bert}
    \end{subfigure}
    \hfill
    \begin{subfigure}[b]{0.24\textwidth}
        \centering
        \includegraphics[width=\textwidth]{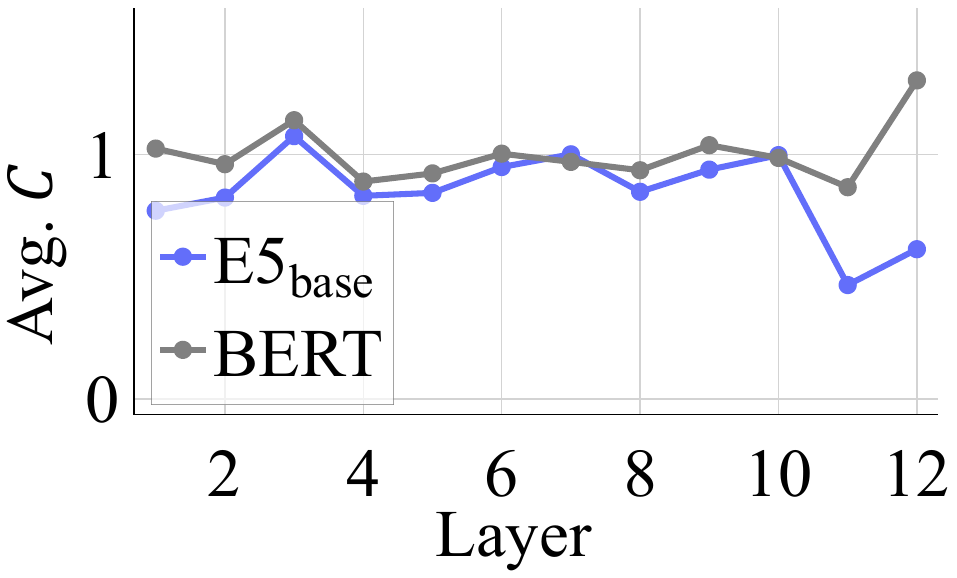}
        \caption{Avg.\ $C$.}
        \label{fig:c_e5base_bert}
    \end{subfigure}
    \hfill
    \begin{subfigure}[b]{0.24\textwidth}
        \centering
        \includegraphics[width=\textwidth]{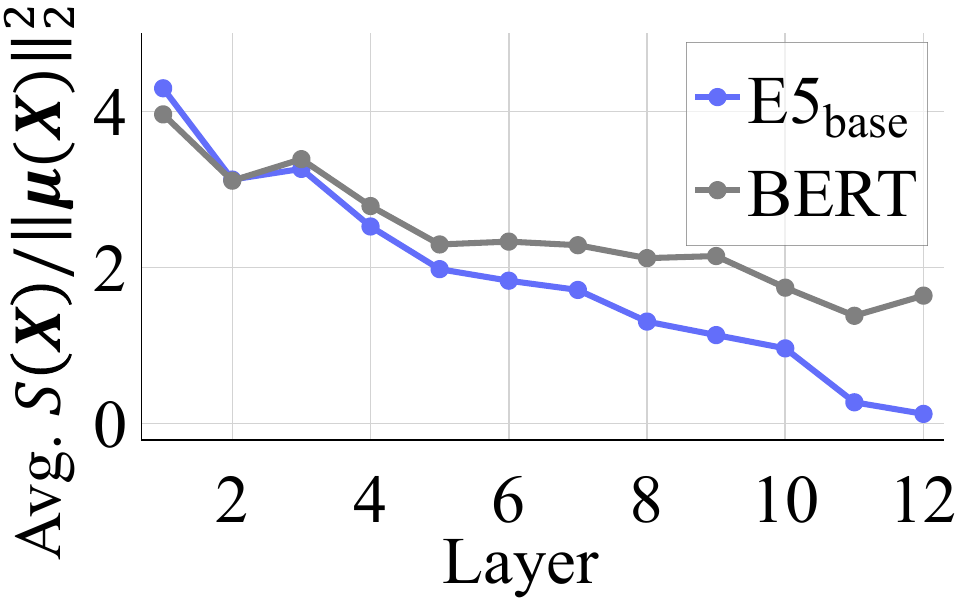}
        \caption{Avg.\ $S(\bm{X})/\|\MU{\bm{X}}\|_2^2$.}
        \label{fig:token_concentration_e5base_bert}
    \end{subfigure}
    \caption{
        Layer-wise analysis of token embedding concentration for BERT and \EFivebase.
        (a) Avg.\ $\lambda$: a smaller value indicates greater concentration of $\bm{Z}$ relative to the input $\bm{H}$.
        (b) Avg.\ $r$: a smaller value indicates a smaller influence of the spread in $\bm{H}$ on the residual output $\bm{Y}=\bm{Z}+\bm{H}$.
        (c) Avg.\ $C$: a smaller value indicates a smaller increase in relative spread of token embeddings from $\bm{Y}$ to $\bm{X}$ through the per-token transformation $\bm{g}$.
        (d) Avg.\ $S(\bm{X})/\|\MU{\bm{X}}\|_2^2$: a smaller value indicates greater concentration of token embeddings within each text.
    }
    \label{fig:how_avoid_collapse_e5base_bert}
\end{figure*}

\begin{figure*}[t]
    \centering
     \begin{subfigure}[b]{0.24\textwidth}
        \centering
        \includegraphics[width=\textwidth]{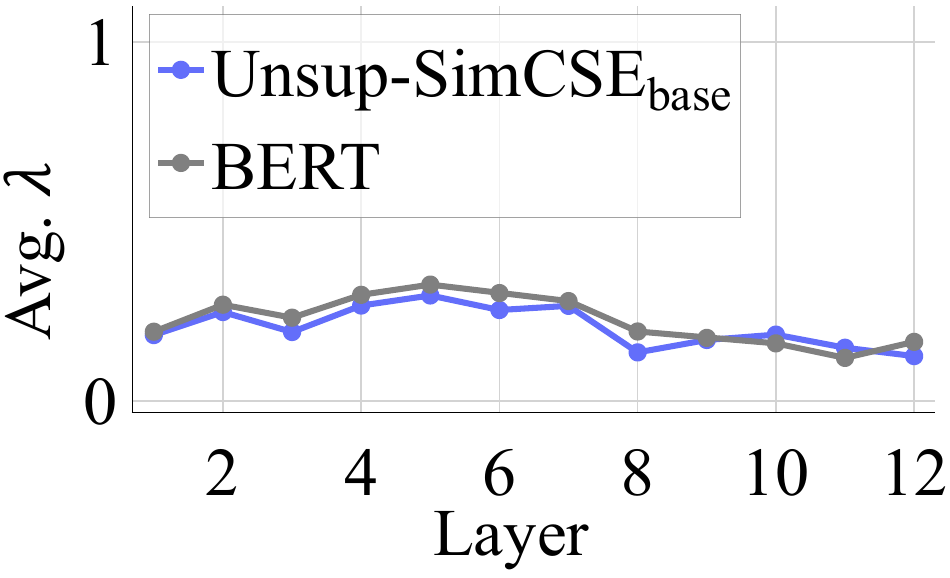}
        \caption{Avg.\ $\lambda$.}
        \label{fig:lambda_simcse_bert}
    \end{subfigure}
    \hfill
    \begin{subfigure}[b]{0.24\textwidth}
        \centering
        \includegraphics[width=\textwidth]{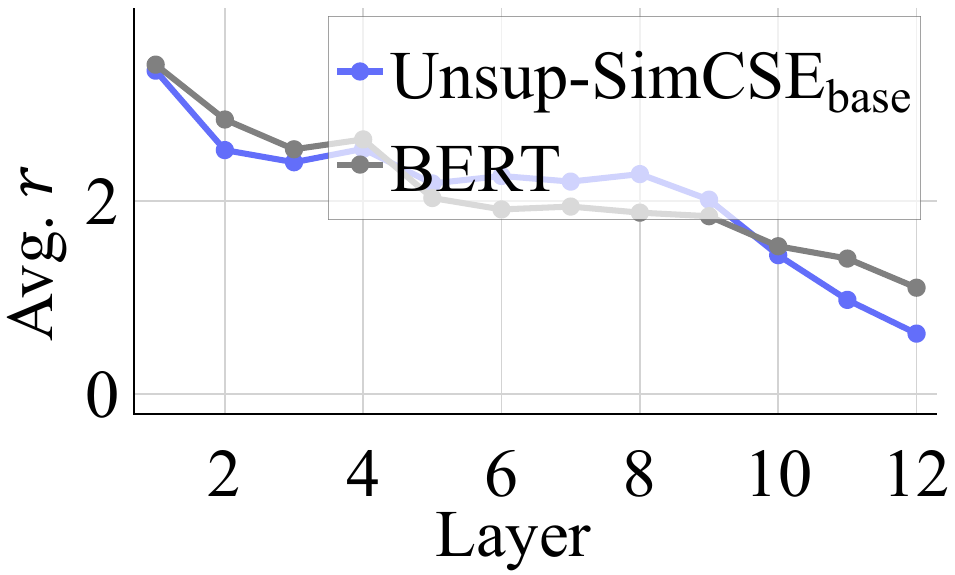}
        \caption{Avg.\ $r$.}
        \label{fig:r_simcse_bert}
    \end{subfigure}
    \hfill
    \begin{subfigure}[b]{0.24\textwidth}
        \centering
        \includegraphics[width=\textwidth]{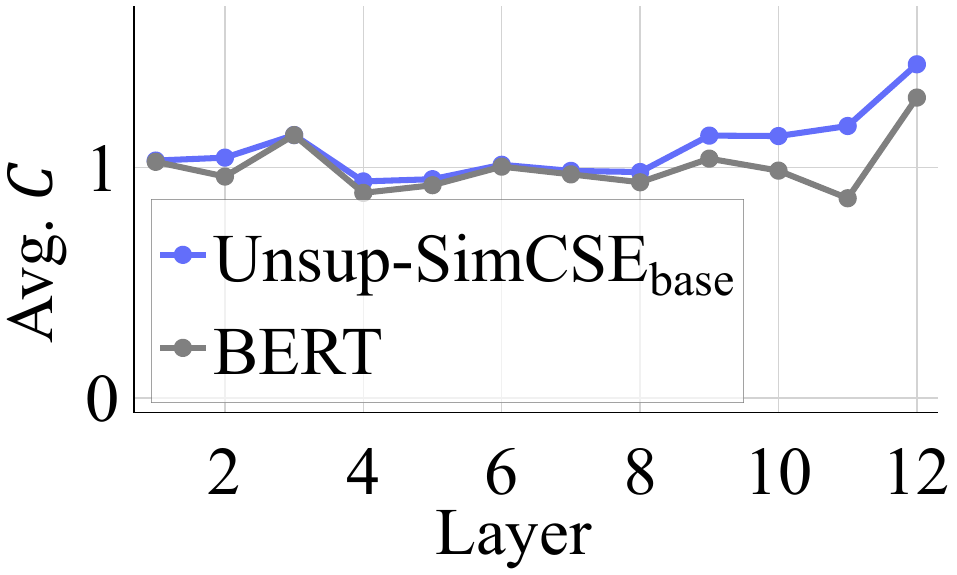}
        \caption{Avg.\ $C$.}
        \label{fig:c_simcse_bert}
    \end{subfigure}
    \hfill
    \begin{subfigure}[b]{0.24\textwidth}
        \centering
        \includegraphics[width=\textwidth]{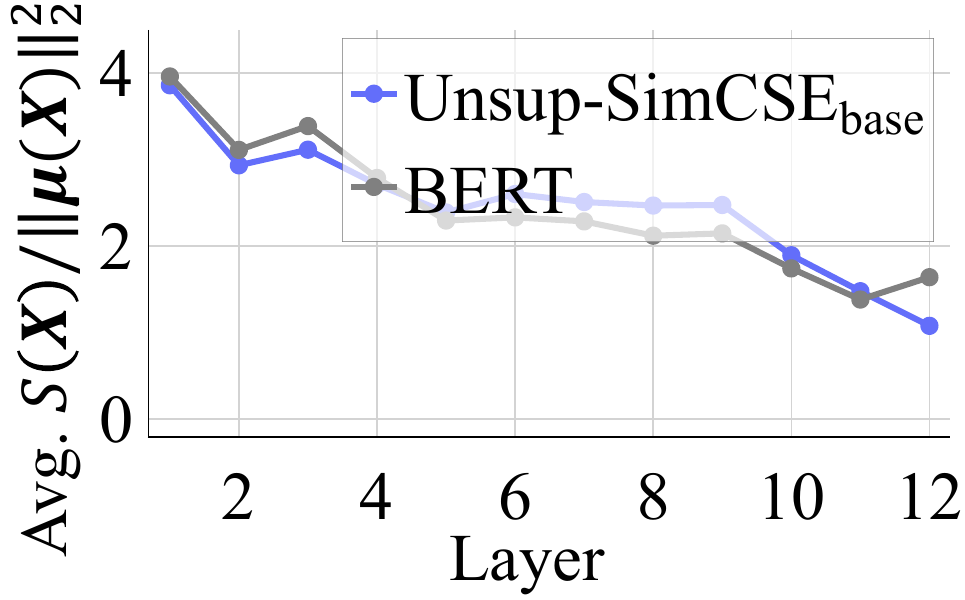}
        \caption{Avg.\ $S(\bm{X})/\|\MU{\bm{X}}\|_2^2$.}
        \label{fig:token_concentration_simcse_bert}
    \end{subfigure}
        \caption{
    Layer-wise analysis of token embedding concentration for BERT and Unsupervised SimCSE.
        (a) Avg.\ $\lambda$: a smaller value indicates greater concentration of $\bm{Z}$ relative to the input $\bm{H}$.
        (b) Avg.\ $r$: a smaller value indicates a smaller influence of the spread in $\bm{H}$ on the residual output $\bm{Y}=\bm{Z}+\bm{H}$.
        (c) Avg.\ $C$: a smaller value indicates a smaller increase in relative spread of token embeddings from $\bm{Y}$ to $\bm{X}$ through the per-token transformation $\bm{g}$.
        (d) Avg.\ $S(\bm{X})/\|\MU{\bm{X}}\|_2^2$: a smaller value indicates greater concentration of token embeddings within each text.
    }
    \label{fig:how_avoid_collapse_simcse_bert}
\end{figure*}

\begin{figure*}[t]
    \centering
     \begin{subfigure}[b]{0.24\textwidth}
        \centering
        \includegraphics[width=\textwidth]{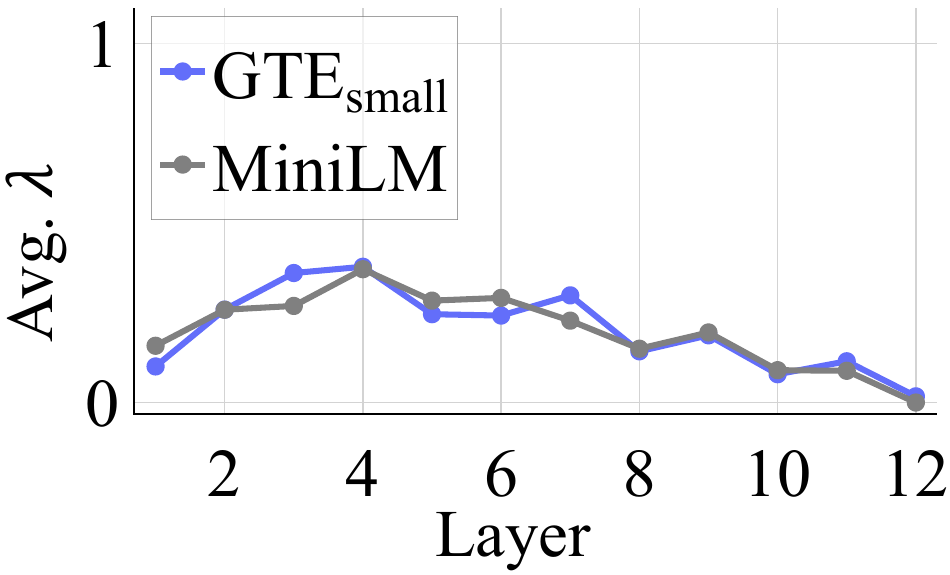}
        \caption{Avg.\ $\lambda$.}
        \label{fig:lambda_gtesmall_minilm}
    \end{subfigure}
    \hfill
    \begin{subfigure}[b]{0.24\textwidth}
        \centering
        \includegraphics[width=\textwidth]{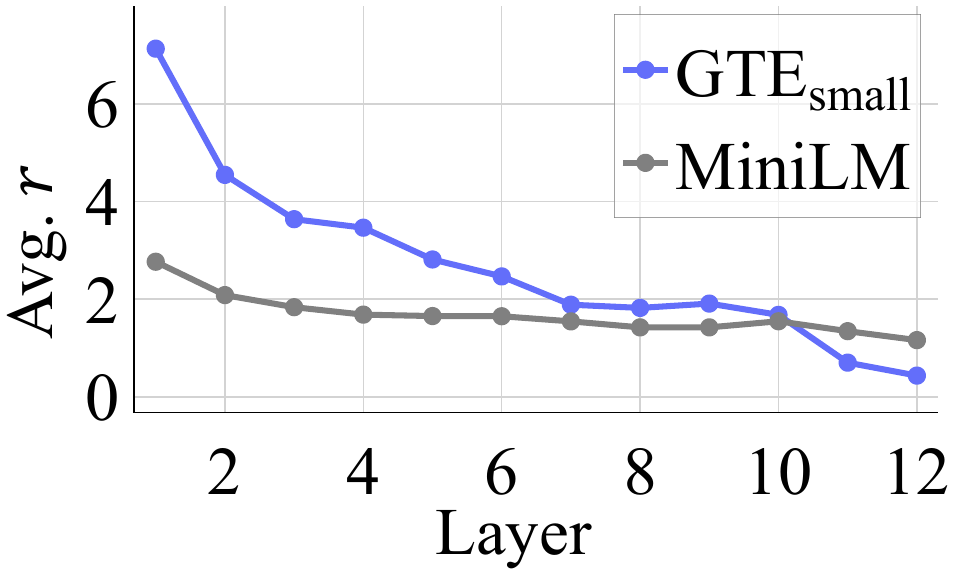}
        \caption{Avg.\ $r$.}
        \label{fig:r_gtesmall_minilm}
    \end{subfigure}
    \hfill
    \begin{subfigure}[b]{0.24\textwidth}
        \centering
        \includegraphics[width=\textwidth]{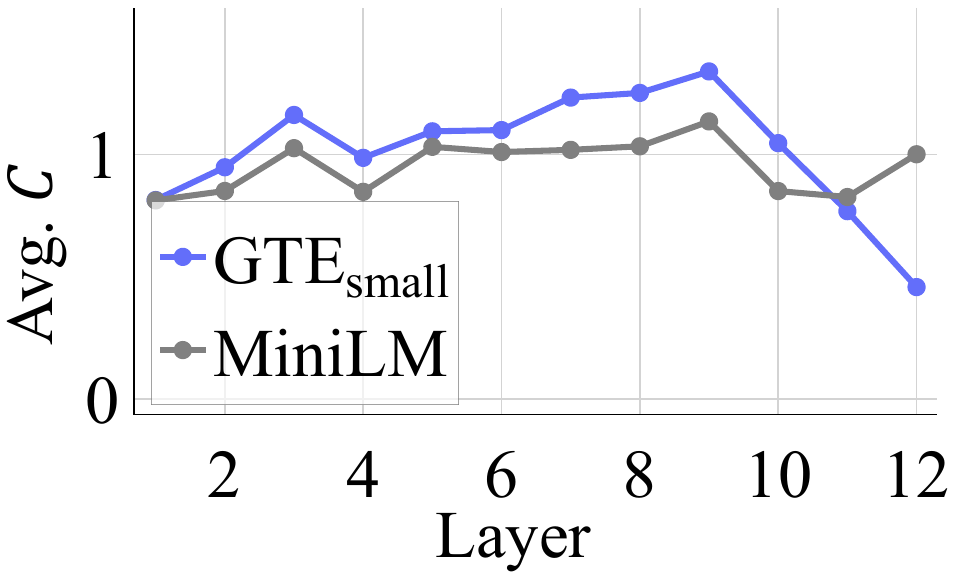}
        \caption{Avg.\ $C$.}
        \label{fig:c_gtesmall_minilm}
    \end{subfigure}
    \hfill
    \begin{subfigure}[b]{0.24\textwidth}
        \centering
        \includegraphics[width=\textwidth]{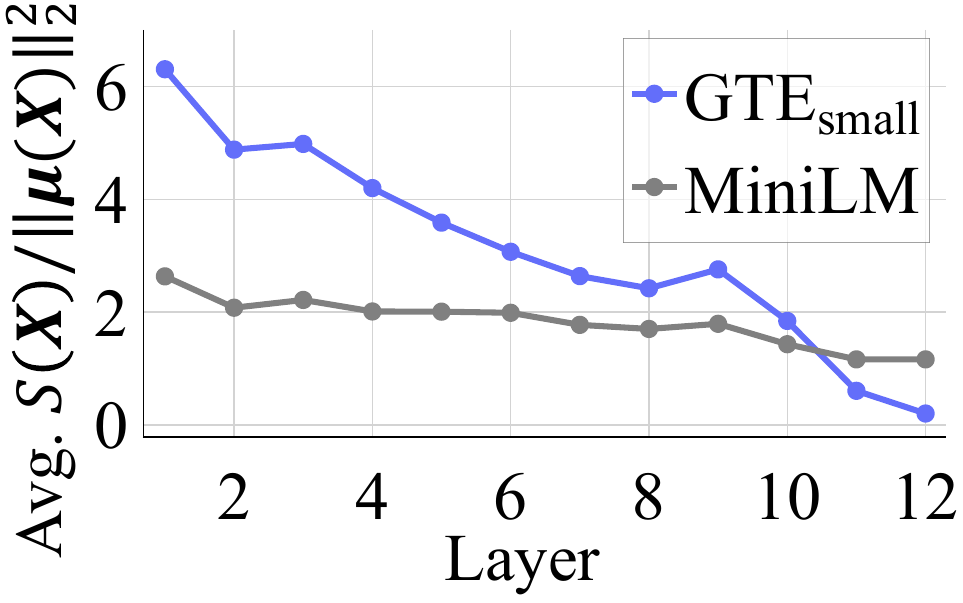}
        \caption{Avg.\ $S(\bm{X})/\|\MU{\bm{X}}\|_2^2$.}
        \label{fig:token_concentration_gtesmall_minilm}
    \end{subfigure}
    \caption{
         Layer-wise analysis of token embedding concentration for MiniLM and \GTEsmall.
        (a) Avg.\ $\lambda$: a smaller value indicates greater concentration of $\bm{Z}$ relative to the input $\bm{H}$.
        (b) Avg.\ $r$: a smaller value indicates a smaller influence of the spread in $\bm{H}$ on the residual output $\bm{Y}=\bm{Z}+\bm{H}$.
        (c) Avg.\ $C$: a smaller value indicates a smaller increase in relative spread of token embeddings from $\bm{Y}$ to $\bm{X}$ through the per-token transformation $\bm{g}$.
        (d) Avg.\ $S(\bm{X})/\|\MU{\bm{X}}\|_2^2$: a smaller value indicates greater concentration of token embeddings within each text.
    }
    \label{fig:how_avoid_collapse_gtesmall_minilm}
\end{figure*}

\begin{figure*}[t]
    \centering
     \begin{subfigure}[b]{0.24\textwidth}
        \centering
        \includegraphics[width=\textwidth]{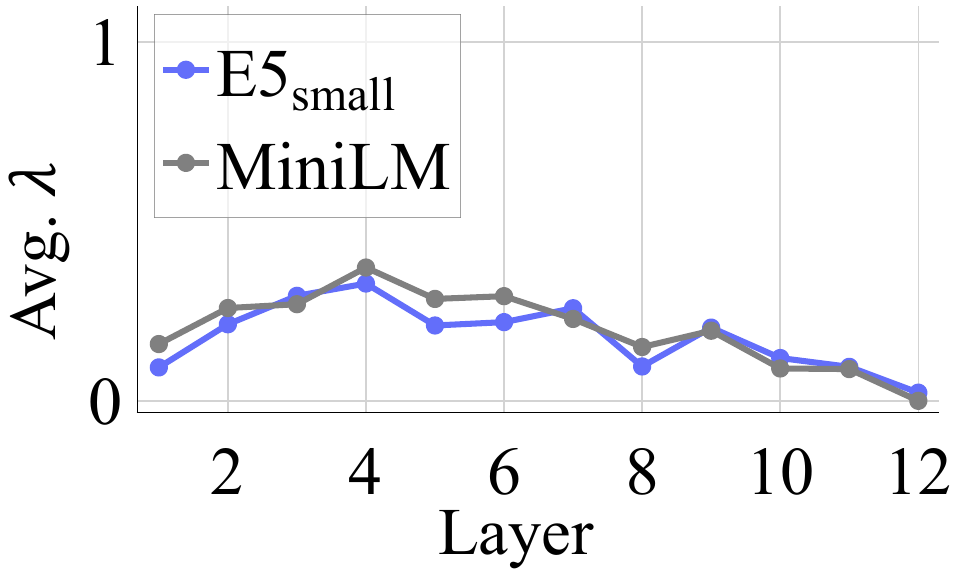}
        \caption{Avg.\ $\lambda$.}
        \label{fig:lambda_e5small_minilm}
    \end{subfigure}
    \hfill
    \begin{subfigure}[b]{0.24\textwidth}
        \centering
        \includegraphics[width=\textwidth]{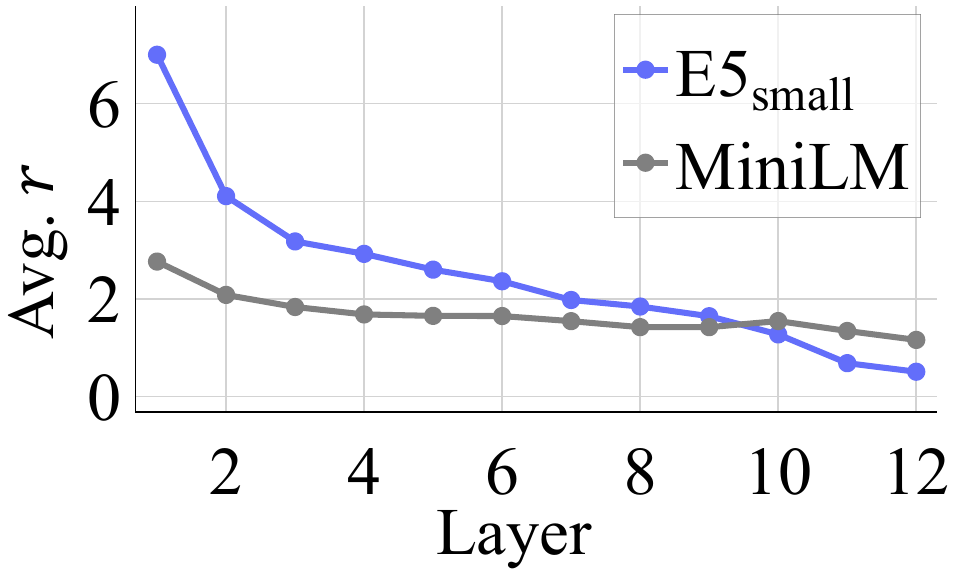}
        \caption{Avg.\ $r$.}
        \label{fig:r_e5small_minilm}
    \end{subfigure}
    \hfill
    \begin{subfigure}[b]{0.24\textwidth}
        \centering
        \includegraphics[width=\textwidth]{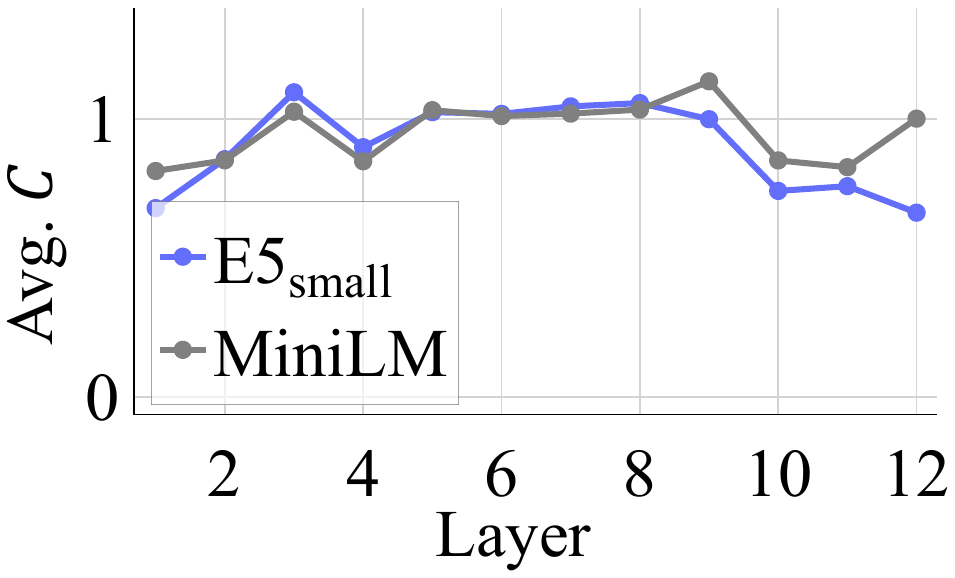}
        \caption{Avg.\ $C$.}
        \label{fig:c_e5small_minilm}
    \end{subfigure}
    \hfill
    \begin{subfigure}[b]{0.24\textwidth}
        \centering
        \includegraphics[width=\textwidth]{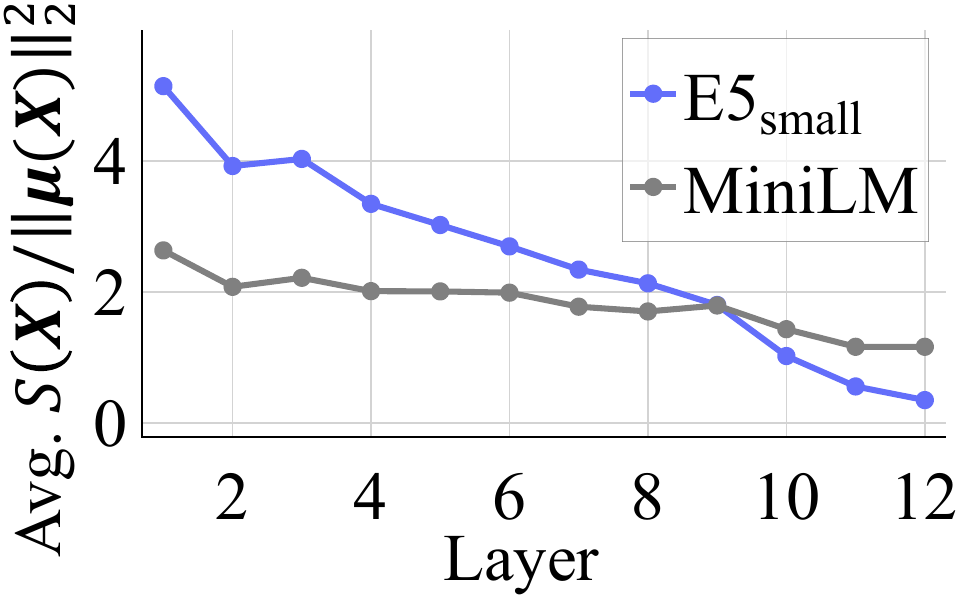}
        \caption{Avg.\ $S(\bm{X})/\|\MU{\bm{X}}\|_2^2$.}
        \label{fig:token_concentration_e5small_minilm}
    \end{subfigure}
    \caption{
       Layer-wise analysis of token embedding concentration for MiniLM and \EFivesmall.
        (a) Avg.\ $\lambda$: a smaller value indicates greater concentration of $\bm{Z}$ relative to the input $\bm{H}$.
        (b) Avg.\ $r$: a smaller value indicates a smaller influence of the spread in $\bm{H}$ on the residual output $\bm{Y}=\bm{Z}+\bm{H}$.
        (c) Avg.\ $C$: a smaller value indicates a smaller increase in relative spread of token embeddings from $\bm{Y}$ to $\bm{X}$ through the per-token transformation $\bm{g}$.
        (d) Avg.\ $S(\bm{X})/\|\MU{\bm{X}}\|_2^2$: a smaller value indicates greater concentration of token embeddings within each text.
    }
    \label{fig:how_avoid_collapse_e5small_minilm}
\end{figure*}

\begin{figure*}[t]
    \centering
     \begin{subfigure}[b]{0.24\textwidth}
        \centering
        \includegraphics[width=\textwidth]{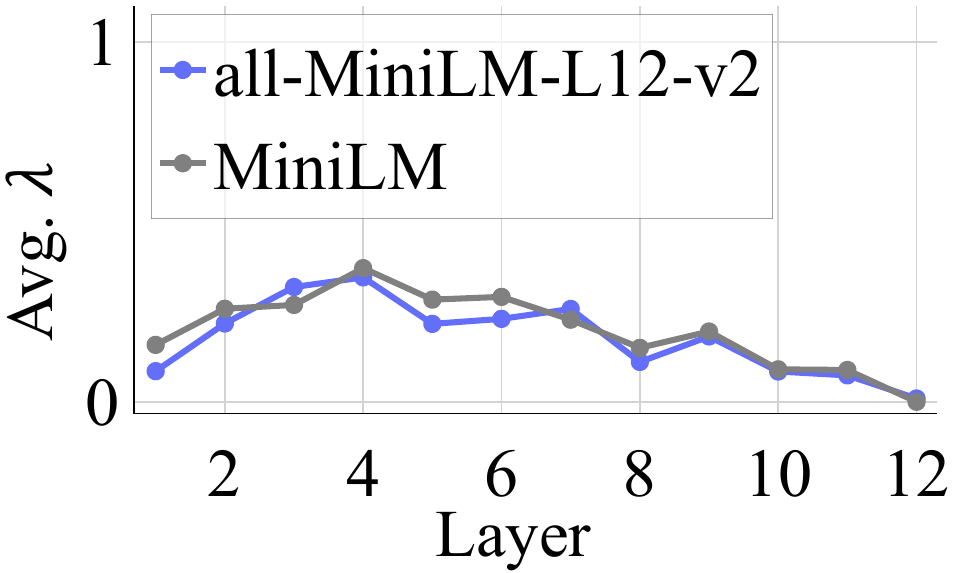}
        \caption{Avg.\ $\lambda$.}
        \label{fig:lambda_allminilm_minilm}
    \end{subfigure}
    \hfill
    \begin{subfigure}[b]{0.24\textwidth}
        \centering
        \includegraphics[width=\textwidth]{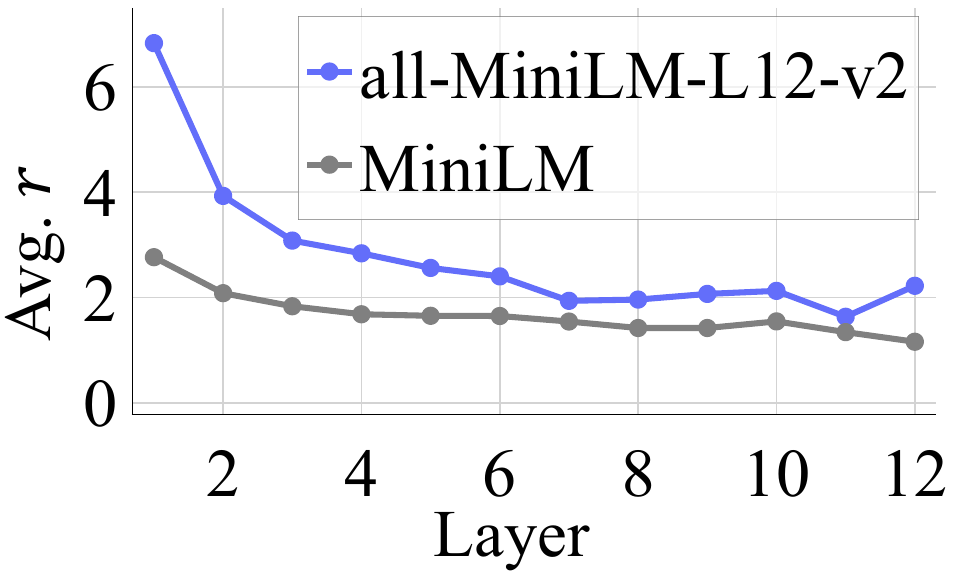}
        \caption{Avg.\ $r$.}
        \label{fig:r_allminilm_minilm}
    \end{subfigure}
    \hfill
    \begin{subfigure}[b]{0.24\textwidth}
        \centering
        \includegraphics[width=\textwidth]{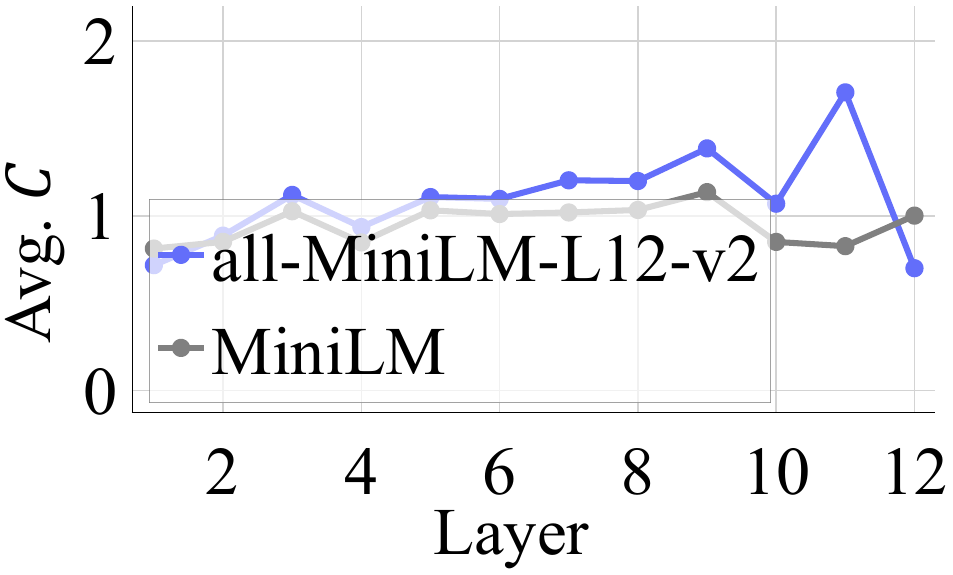}
        \caption{Avg.\ $C$.}
        \label{fig:c_allminilm_minilm}
    \end{subfigure}
    \hfill
    \begin{subfigure}[b]{0.24\textwidth}
        \centering
        \includegraphics[width=\textwidth]{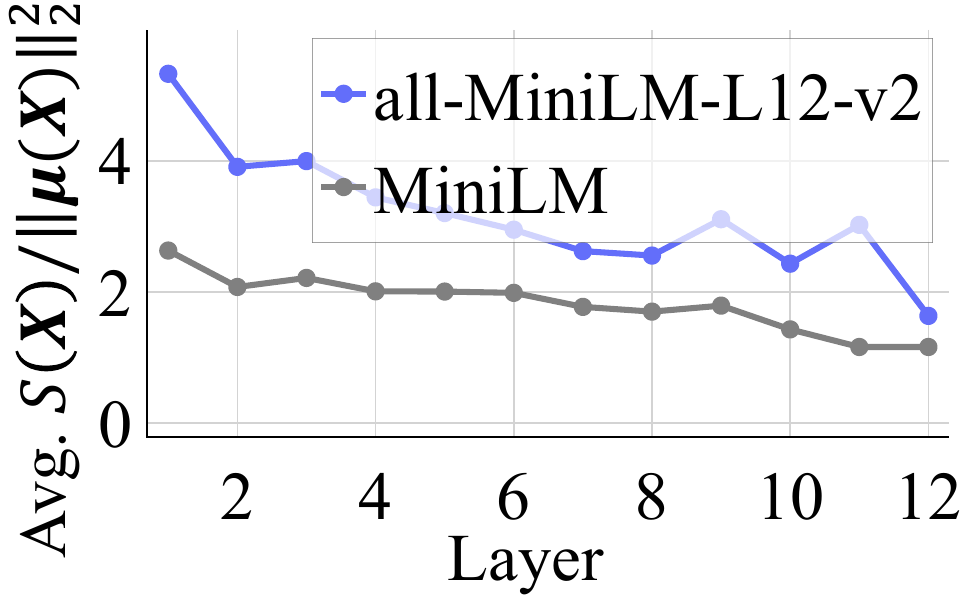}
        \caption{Avg.\ $S(\bm{X})/\|\MU{\bm{X}}\|_2^2$.}
        \label{fig:token_concentration_allminilm_minilm}
    \end{subfigure}
    \caption{
        Layer-wise analysis of token embedding concentration for MiniLM and all-MiniLM-L12-v2.
        (a) Avg.\ $\lambda$: a smaller value indicates greater concentration of $\bm{Z}$ relative to the input $\bm{H}$.
        (b) Avg.\ $r$: a smaller value indicates a smaller influence of the spread in $\bm{H}$ on the residual output $\bm{Y}=\bm{Z}+\bm{H}$.
        (c) Avg.\ $C$: a smaller value indicates a smaller increase in relative spread of token embeddings from $\bm{Y}$ to $\bm{X}$ through the per-token transformation $\bm{g}$.
        (d) Avg.\ $S(\bm{X})/\|\MU{\bm{X}}\|_2^2$: a smaller value indicates greater concentration of token embeddings within each text.
    }
    \label{fig:how_avoid_collapse_allminilm_minilm}
\end{figure*}

We provide additional results for the layer-wise analysis of token embedding concentration described in \S~\ref{sec:how_avoid_collapse}.
In addition to the results for BERT and \GTEbase reported in \S~\ref{sec:how_avoid_collapse}, Figures~\ref{fig:how_avoid_collapse_e5base_bert}--\ref{fig:how_avoid_collapse_allminilm_minilm} show the layer-wise averages of $\lambda$, $r$, $C$, and $S(\bm{X})/\|\MU{\bm{X}}\|_2^2$ for the remaining model pairs: \EFivebase and BERT (Figure~\ref{fig:how_avoid_collapse_e5base_bert}), Unsupervised SimCSE and BERT (Figure~\ref{fig:how_avoid_collapse_simcse_bert}), \GTEsmall and MiniLM (Figure~\ref{fig:how_avoid_collapse_gtesmall_minilm}), \EFivesmall and MiniLM (Figure~\ref{fig:how_avoid_collapse_e5small_minilm}), and all-MiniLM-L12-v2 and MiniLM (Figure~\ref{fig:how_avoid_collapse_allminilm_minilm}).
These results tend to show similar trends to those observed for BERT and \GTEbase in \S~\ref{sec:how_avoid_collapse}.

\subsection{Within-Text Token Embedding Concentration}\label{sec:appendix_avg_cos}
\begin{figure}[t]
    \centering
    \includegraphics[width=0.7\linewidth]{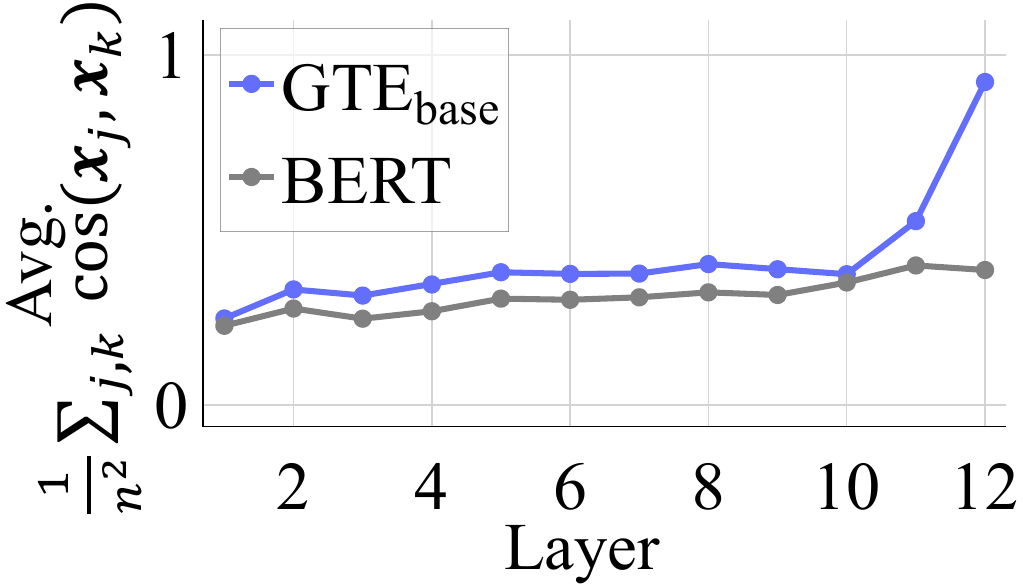}
    \caption{
    Layer-wise average of $\frac{1}{n^2}\sum_{j,k}\cos(\bm{x}_j, \bm{x}_k)$ for BERT and \GTEbase on the Wikipedia dataset.
    }
    \label{fig:avg_cos_gtebase_bert}
\end{figure}

\begin{figure}[t]
    \centering
    \includegraphics[width=0.7\linewidth]{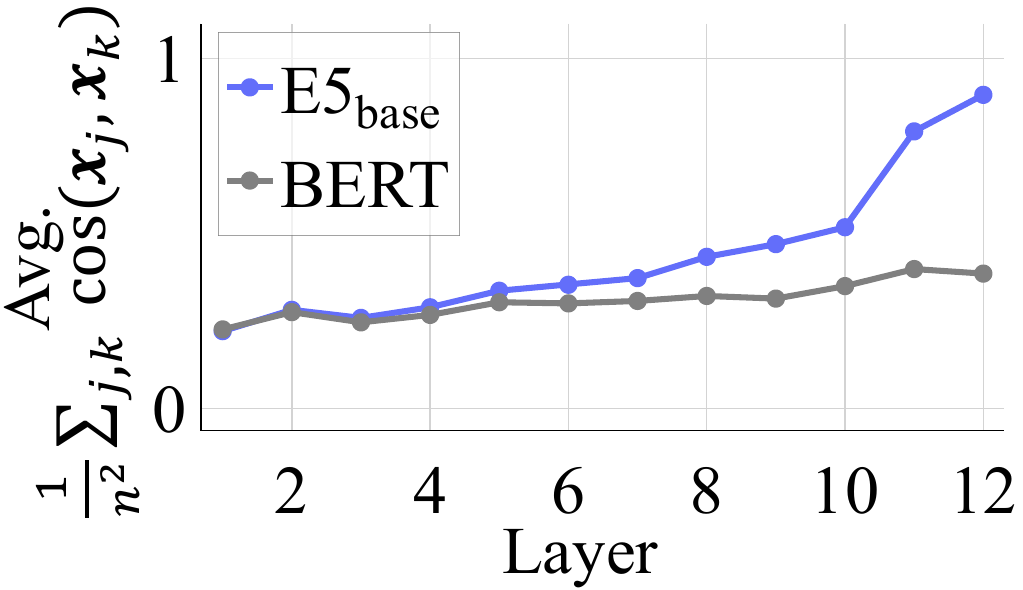}
    \caption{
    Layer-wise average of $\frac{1}{n^2}\sum_{j,k}\cos(\bm{x}_j, \bm{x}_k)$ for BERT and \EFivebase on the Wikipedia dataset.
    }
    \label{fig:avg_cos_e5base_bert}
\end{figure}

\begin{figure}[t]
    \centering
    \includegraphics[width=0.7\linewidth]{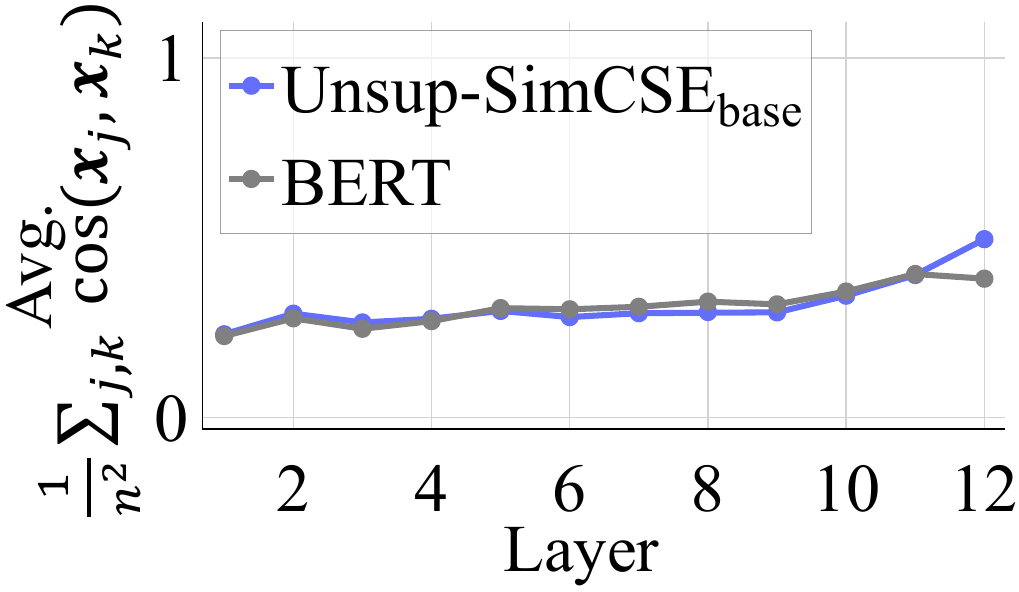}
    \caption{
    Layer-wise average of $\frac{1}{n^2}\sum_{j,k}\cos(\bm{x}_j, \bm{x}_k)$ for BERT and Unsupervised SimCSE on the Wikipedia dataset.
    }
    \label{fig:avg_cos_simcse_bert}
\end{figure}

\begin{figure}[t]
    \centering
    \includegraphics[width=0.7\linewidth]{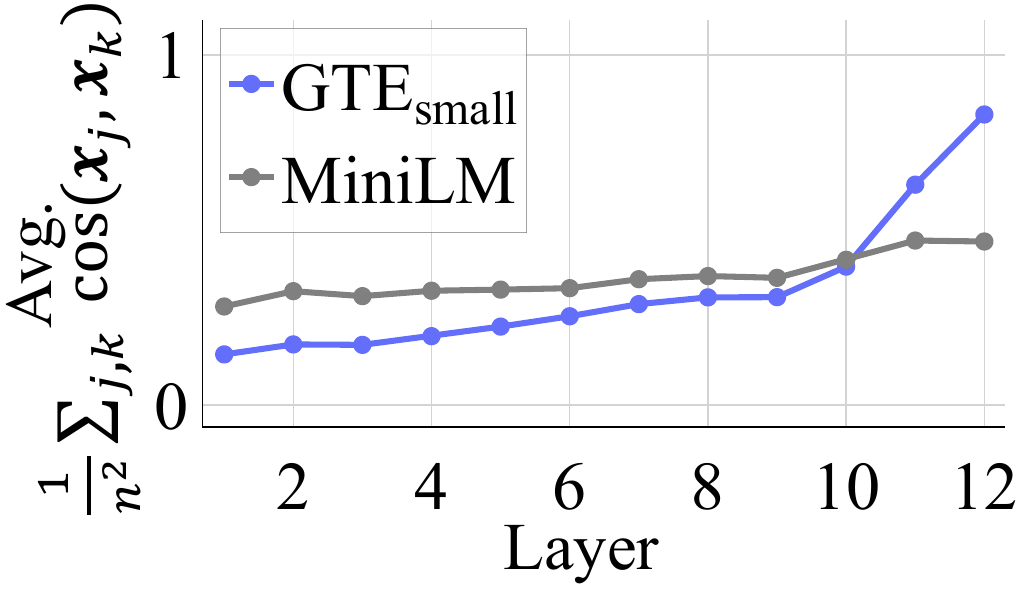}
    \caption{
    Layer-wise average of $\frac{1}{n^2}\sum_{j,k}\cos(\bm{x}_j, \bm{x}_k)$ for MiniLM and \GTEsmall on the Wikipedia dataset.
    }
    \label{fig:avg_cos_gtesmall_minilm}
\end{figure}

\begin{figure}[t]
    \centering
    \includegraphics[width=0.7\linewidth]{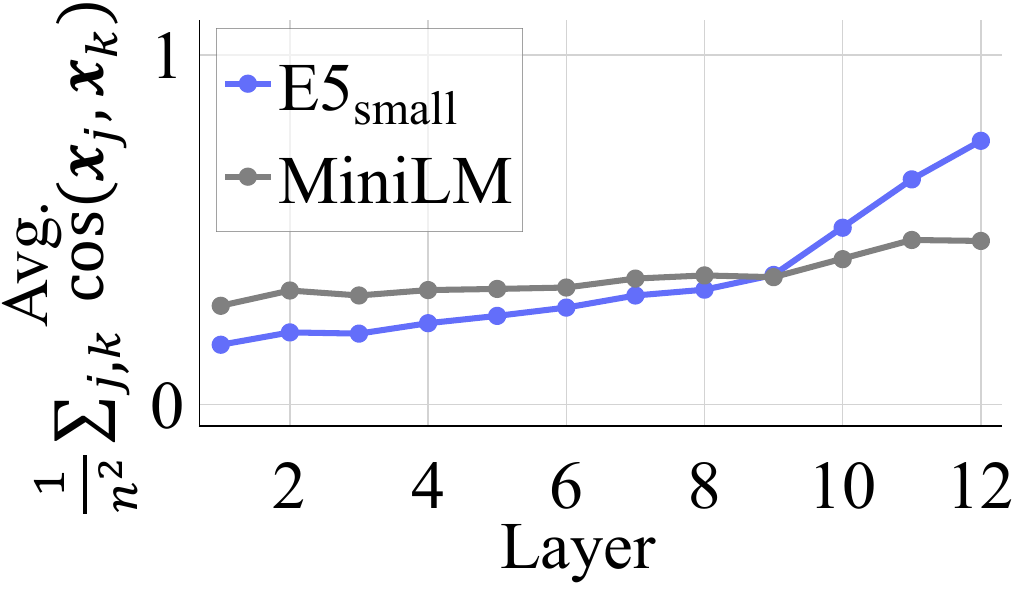}
    \caption{
    Layer-wise average of $\frac{1}{n^2}\sum_{j,k}\cos(\bm{x}_j, \bm{x}_k)$ for MiniLM and \EFivesmall on the Wikipedia dataset.
    }
    \label{fig:avg_cos_e5small_minilm}
\end{figure}

\begin{figure}[t]
    \centering
    \includegraphics[width=0.7\linewidth]{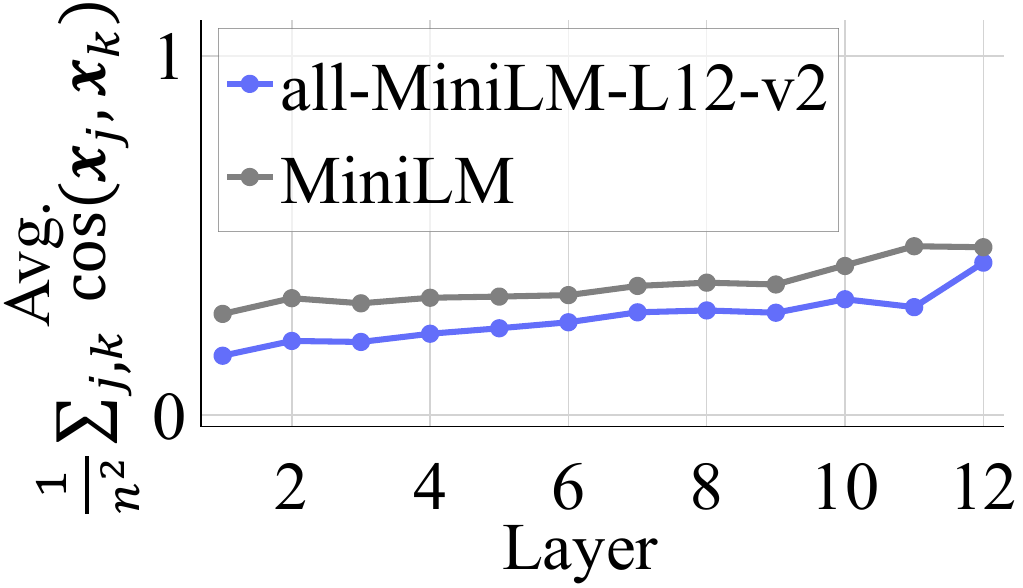}
    \caption{
    Layer-wise average of $\frac{1}{n^2}\sum_{j,k}\cos(\bm{x}_j, \bm{x}_k)$ for MiniLM and all-MiniLM-L12-v2 on the Wikipedia dataset.
    }
    \label{fig:avg_cos_allminilm_minilm}
\end{figure}

In \S~\ref{sec:how_avoid_collapse}, we connected our findings to the anisotropy of token embeddings within each text reported by \citet{Xiao2023-vu}.
As anisotropy is commonly measured using cosine similarity~\citep{Xiao2023-vu}, we quantify within-text token embedding concentration via the layer-wise average of $\frac{1}{n^2}\sum_{j,k}\cos(\bm{x}_j, \bm{x}_k)$, where $\bm{X} = [\bm{x}_1, \ldots, \bm{x}_n]$ denotes the token embeddings of a text, and each average is taken over all texts in the dataset.\footnote{Note that this quantity is not identical to the anisotropy measure used by \citet{Xiao2023-vu}.}
A higher value indicates greater concentration, consistent with higher within-text anisotropy as described by \citet{Xiao2023-vu}.
Figures~\ref{fig:avg_cos_gtebase_bert}--\ref{fig:avg_cos_allminilm_minilm} show these results for the following model pairs:
\GTEbase and BERT (Figure~\ref{fig:avg_cos_gtebase_bert}),
\EFivebase and BERT (Figure~\ref{fig:avg_cos_e5base_bert}),
Unsupervised SimCSE and BERT (Figure~\ref{fig:avg_cos_simcse_bert}),
\GTEsmall and MiniLM (Figure~\ref{fig:avg_cos_gtesmall_minilm}),
\EFivesmall and MiniLM (Figure~\ref{fig:avg_cos_e5small_minilm}), and
all-MiniLM-L12-v2 and MiniLM (Figure~\ref{fig:avg_cos_allminilm_minilm}).
Fine-tuned text encoders tend to show higher within-text average cosine similarity than their corresponding backbone models, particularly in the later layers.
This trend is consistent with the lower \OurTool and smaller $S(\bm{X})/\|\MU{\bm{X}}\|_2^2$ observed in fine-tuned text encoders (\S~\ref{sec:how_avoid_collapse}), supporting the connection between within-text token embedding concentration and robustness against collapse by mean pooling.

\section{Analysis of \texorpdfstring{$d_\mu$}{d\_mu} and \texorpdfstring{$d_\Sigma$}{d\_Sigma}}\label{sec:appendix_scatter}
We examine the components $d_\mu$ and $d_\Sigma$ that constitute \OurTool to understand the variation in \OurTool across model pairs observed in \S~\ref{sec:experiment}.

\paragraph{Observations}
\begin{figure*}[t]
    \centering
    \includegraphics[width=\linewidth]{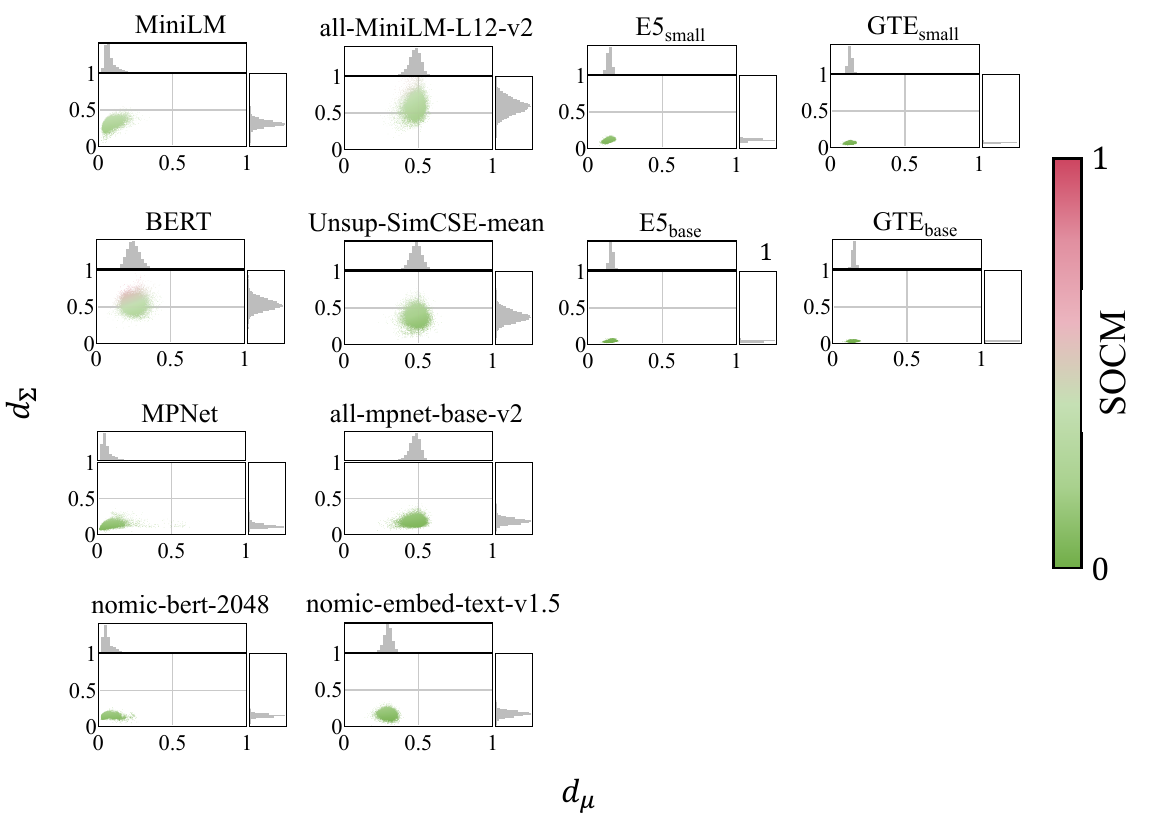}
    \caption{
    Scatter plots of $d_\mu$ and $d_\Sigma$ for the models examined in \S~\ref{sec:experiment} on Wikipedia.
    Each point represents a text pair, colored by \OurTool.
    The top and right marginal histograms show the distributions of $d_\mu$ and $d_\Sigma$.
    }
    \label{fig:scatter_socm}
\end{figure*}

Figure~\ref{fig:scatter_socm} shows scatter plots of $d_\mu$ and $d_\Sigma$ for all examined models on the Wikipedia dataset.
Across all model pairs, fine-tuned text encoders tend to exhibit small $d_\Sigma$ values.
\paragraph{Connection to Token Embedding Concentration}
This tendency is consistent with the results of \S~\ref{sec:how_avoid_collapse}, which showed that fine-tuned text encoders tend to concentrate token embeddings around their within-text mean.
When token embeddings concentrate, their within-text covariance $\SIGMA{\bm{X}}$ becomes small.
This directly suppresses $d_\Sigma$: since $d_\Sigma$ is the scaled Bures-Wasserstein distance between $\SIGMA{\bm{X}_1}$ and $\SIGMA{\bm{X}_2}$, it satisfies $d_\Sigma \leq \bigl(\mathrm{tr}(\SIGMA{\bm{X}_1}) + \mathrm{tr}(\SIGMA{\bm{X}_2})\bigr)/4$.
Therefore, when both $\SIGMA{\bm{X}_1}$ and $\SIGMA{\bm{X}_2}$ are small, $d_\Sigma$ is bounded to be small as well, regardless of how different the two covariance matrices are from each other.

\section{Trace Bound under Normalization} \label{sec:appendix_trace_bound}
In this section, we prove that the assumption $\mathrm{tr}(\SIGMA{\bm{X}_i}) \leq 2$ in \S~\ref{subsubsec:socm_definition} holds under the normalization procedure described in \S~\ref{sec:experiment} and certain assumptions about the model architecture.

\subsection{Setup and Notation}
We consider token embeddings output from a LayerNorm layer~\cite{Devlin2019-mb}.
Let $\bm{y}_{i,j} \in \mathbb{R}^d$ denote the hidden state before LayerNorm for the $j$-th token in text $i$.
The $k$-th dimension of the token embedding after LayerNorm is given by:
\begin{equation}
    x_{i,j,k} = \gamma_k \cdot \frac{y_{i,j,k} - m_{i,j}}{s_{i,j}} + \beta_k,
\end{equation}
where $\gamma_k, \beta_k \in \mathbb{R}$ are learnable parameters for dimension $k$, and $m_{i,j}$ and $s_{i,j}$ are the mean and standard deviation computed across dimensions:
\begin{align}
    m_{i,j} &= \frac{1}{d}\sum_{k=1}^d y_{i,j,k}, \\
    s_{i,j} &= \sqrt{\frac{1}{d}\sum_{k=1}^d (y_{i,j,k} - m_{i,j})^2}.
\end{align}
Let $\bm{x}_{i,j} = [x_{i,j,1}, \ldots, x_{i,j,d}]^\top \in \mathbb{R}^d$ denote the token embedding after LayerNorm.
The token embedding list before normalization is $\bm{X}_i = [\bm{x}_{i,1}, \ldots, \bm{x}_{i,n_i}]$ with $\MU{\bm{X}_i} = \frac{1}{n_i}\sum_{j=1}^{n_i}\bm{x}_{i,j}$.
After normalization by the mean norm (as described in \S~\ref{sec:experiment}), we obtain the normalized token embedding list $\bm{X}_i^\text{norm} = [\bm{x}_{i,1}^\text{norm}, \ldots, \bm{x}_{i,n_i}^\text{norm}]$, where:
\begin{equation}
    \bm{x}_{i,j}^\text{norm} = \frac{\bm{x}_{i,j}}{\|\MU{\bm{X}_i}\|_2}.
\end{equation}

\subsection{Assumptions}
We make the following assumptions:
\paragraph{Assumption 1}
LayerNorm parameters are shared across all dimensions, i.e., $\gamma_1 = \cdots = \gamma_d = \gamma$ and $\beta_1 = \cdots = \beta_d = \beta$.
This assumption reflects the initialization of LayerNorm parameters in typical Transformer models~\cite{Devlin2019-mb}, where $\gamma_1 = \cdots = \gamma_d = 1$ and $\beta_1 = \cdots = \beta_d = 0$.

\paragraph{Assumption 2}
Token embeddings within the same text exhibit sufficient similarity.
Specifically, the expected cosine similarity between token embeddings within the same text satisfies $\mathbb{E}_{j<k}[\cos(\bm{x}_{i,j}, \bm{x}_{i,k})] \geq 1/3$.
This assumption is justified by empirical observations that contextualized embeddings within the same text are anisotropic~\cite{Ethayarajh2019-yw}.

\subsection{Proof}
Under Assumptions 1 and 2, we show that $\mathrm{tr}(\SIGMA{\bm{X}_i^\text{norm}}) \leq 2$ holds for the normalized token embedding list $\bm{X}_i^\text{norm}$.

\paragraph{Relating normalized and unnormalized covariance matrices}
We first establish that:
\begin{equation}
    \mathrm{tr}(\SIGMA{\bm{X}_i^\text{norm}}) = \frac{1}{\|\MU{\bm{X}_i}\|_2^2}\mathrm{tr}(\SIGMA{\bm{X}_i}),
\end{equation}
where $\SIGMA{\bm{X}_i} = \frac{1}{n_i}\sum_{j=1}^{n_i}(\bm{x}_{i,j} - \MU{\bm{X}_i})(\bm{x}_{i,j} - \MU{\bm{X}_i})^\top$.
This follows from the definition $\bm{x}_{i,j}^\text{norm} = \frac{\bm{x}_{i,j}}{\|\MU{\bm{X}_i}\|_2}$ and the properties of the trace operator.

\paragraph{Computing the trace of unnormalized covariance}
Under Assumption 1, the norm of each token embedding after LayerNorm is constant:
\begin{equation}
    \|\bm{x}_{i,j}\|_2^2 = \sum_{k=1}^d x_{i,j,k}^2 = d(\gamma^2 + \beta^2).
\end{equation}
Using this fact and the relation $\mathrm{tr}(\SIGMA{\bm{X}_i}) = \frac{1}{n_i}\sum_{j=1}^{n_i}\|\bm{x}_{i,j}\|_2^2 - \|\MU{\bm{X}_i}\|_2^2$, we obtain:
\begin{equation}
    \mathrm{tr}(\SIGMA{\bm{X}_i}) = d(\gamma^2 + \beta^2) - \|\MU{\bm{X}_i}\|_2^2.
\end{equation}

\paragraph{Computing the squared norm of the mean}
The squared norm of the mean can be expressed as:
\begin{align}
    &\|\MU{\bm{X}_i}\|_2^2 
    = \frac{1}{n_i^2}\sum_{j=1}^{n_i}\sum_{k=1}^{n_i}\bm{x}_{i,j}^\top\bm{x}_{i,k} \\
    &= \frac{1}{n_i^2}\left(\sum_{j=1}^{n_i}\|\bm{x}_{i,j}\|_2^2 + 2\sum_{j<k}\bm{x}_{i,j}^\top\bm{x}_{i,k}\right) \\
    &= \frac{1}{n_i^2}\Bigl(n_i d(\gamma^2 + \beta^2) \\ \nonumber
    &\qquad + 2\sum_{j<k}d(\gamma^2 + \beta^2)\cos(\bm{x}_{i,j}, \bm{x}_{i,k})\Bigr) \\
    &= \frac{d(\gamma^2 + \beta^2)}{n_i}\Bigl(1 \\ \nonumber
    &\qquad + (n_i - 1)\mathbb{E}_{j<k}[\cos(\bm{x}_{i,j}, \bm{x}_{i,k})]\Bigr).
\end{align}

\paragraph{Deriving the expression for the trace of normalized covariance}
Combining the above results:
\begin{align}
    &\mathrm{tr}(\SIGMA{\bm{X}_i^\text{norm}}) 
    = \frac{d(\gamma^2 + \beta^2) - \|\MU{\bm{X}_i}\|_2^2}{\|\MU{\bm{X}_i}\|_2^2} \\
    &= \frac{1}{\|\MU{\bm{X}_i}\|_2^2}d(\gamma^2 + \beta^2) - 1 \\
    &= \frac{d(\gamma^2 + \beta^2)}{\frac{d(\gamma^2 + \beta^2)}{n_i}\left(1 + (n_i - 1)\mathbb{E}_{j<k}[\cos(\bm{x}_{i,j}, \bm{x}_{i,k})]\right)}\\ \nonumber 
    &\qquad- 1 \\
    &= \frac{n_i}{1 + (n_i - 1)\mathbb{E}_{j<k}[\cos(\bm{x}_{i,j}, \bm{x}_{i,k})]} - 1 \\
    &= \frac{(n_i - 1)(1 - \mathbb{E}_{j<k}[\cos(\bm{x}_{i,j}, \bm{x}_{i,k})])}{1 + (n_i - 1)\mathbb{E}_{j<k}[\cos(\bm{x}_{i,j}, \bm{x}_{i,k})]}.
\end{align}

\paragraph{Bounding the trace}
This expression is monotonically decreasing in $\mathbb{E}_{j<k}[\cos(\bm{x}_{i,j}, \bm{x}_{i,k})]$.
To verify this, we compute:
\begin{align}
    &\frac{\partial}{\partial \mathbb{E}_{j<k}[\cos(\bm{x}_{i,j}, \bm{x}_{i,k})]}\mathrm{tr}(\SIGMA{\bm{X}_i^\text{norm}}) \\ \nonumber
    &= -\frac{n_i}{\left(1 + (n_i - 1)\mathbb{E}_{j<k}[\cos(\bm{x}_{i,j}, \bm{x}_{i,k})]\right)^2} < 0.
\end{align}
Under Assumption 2, when $\mathbb{E}_{j<k}[\cos(\bm{x}_{i,j}, \bm{x}_{i,k})] = 1/3$, we have:
\begin{equation}
    \mathrm{tr}(\SIGMA{\bm{X}_i^\text{norm}}) = \frac{(n_i - 1) \cdot \frac{2}{3}}{1 + (n_i - 1) \cdot \frac{1}{3}} = \frac{2(n_i - 1)}{n_i + 2}.
\end{equation}
This function is monotonically increasing in $n_i$:
\begin{equation}
    \frac{\partial}{\partial n_i}\frac{2(n_i - 1)}{n_i + 2} = \frac{6}{(n_i + 2)^2} > 0.
\end{equation}
Furthermore, this function converges to 2 as $n_i \to \infty$:
\begin{equation}
    \lim_{n_i \to \infty}\frac{2(n_i - 1)}{n_i + 2} = 2.
\end{equation}
Therefore, $\mathrm{tr}(\SIGMA{\bm{X}_i^\text{norm}}) \leq 2$ holds under Assumptions 1 and 2.

\section{Computational Resources} \label{sec:appendix_computational_resources}
All experiments in this paper were conducted using a single NVIDIA RTX 6000 Ada graphics card.
In \S~\ref{sec:experiment}, computing \OurTool values for all text pairs required approximately 2 hours per model.
The analyses with MTEB (eng, v2) in \S~\ref{sec:correlation_downstream_task} required approximately 6 hours per model.

\section{Use of AI Assistants}
In preparing this paper, we utilized AI assistants (Claude, ChatGPT) to support various aspects of the writing and implementation process.
These tools were employed for tasks such as code debugging, language polishing, formatting assistance, and generating visualizations.
However, all research ideas, methodological designs, experimental analyses, and scientific interpretations presented in this work are entirely our own.
The AI assistants served solely as technical aids and did not contribute to the conceptual or intellectual content of this research.
\end{document}